%% file: paper.tex
\documentclass{article}

\usepackage{microtype}
\usepackage{graphicx}
\usepackage{subcaption}
\usepackage{booktabs} 
\usepackage{longtable}
\usepackage{caption}

\usepackage{hyperref}

\usepackage[preprint]{icml2026}

\usepackage{amsmath}
\usepackage{amssymb}
\usepackage{mathtools}
\usepackage{amsthm}
\usepackage{multirow}
\usepackage[table]{xcolor}
\usepackage{pifont}

\usepackage[capitalize,noabbrev]{cleveref}

\theoremstyle{plain}

\theoremstyle{definition}

\theoremstyle{remark}

\usepackage[textsize=tiny]{todonotes}

\usepackage[ruled,noline,algo2e]{algorithm2e}
\usepackage{setspace}  

\usepackage{tcolorbox}
\definecolor{bg-gray}{RGB}{248, 248, 248} 
\definecolor{hl-pink}{RGB}{255, 160, 180} 
\definecolor{hl-yellow}{RGB}{240, 200, 100}

\definecolor{commentcolor}{RGB}{110,154,155}   
\definecolor{kwcolor}{RGB}{202, 33, 114}   
\newcommand{\PyComment}[1]{\ttfamily\textcolor{commentcolor}{\# #1}}  
\newcommand{\PyCode}[1]{\ttfamily\textcolor{black}{#1}} 
\newcommand{\PyKW}[1]{\ttfamily\textcolor{kwcolor}{#1}}

\newtcolorbox{takeaway}{
    colback=bg-gray,
    colframe=black,
    coltext=black,
    boxrule=1.0pt,
    arc=3pt,
    auto outer arc,
    boxsep=0pt,
    left=6pt,
    right=6pt,
    top=5pt,
    bottom=5pt,
    before skip=4pt,
    after skip=4pt,
    fontupper=\small,
    sharp corners=northwest,
}

\icmltitlerunning{\textsc{Eidos}: Latent-Space Predictive Learning for Time Series Foundation Models}

\newcommand{\modelname}{\textsc{Eidos}\xspace}

\usepackage{soul}

\begin{document}

\twocolumn[
  \icmltitle{\textsc{Eidos}: Latent-Space Predictive Learning for Time Series Foundation Models}



  \icmlsetsymbol{equal}{*}

  \begin{icmlauthorlist}
    \icmlauthor{Xinxing Zhou}{nk}
    \icmlauthor{Qingren Yao}{tue}
    \icmlauthor{Yiji Zhao}{ynu}
    \icmlauthor{Chenghao Liu}{dg}
    \icmlauthor{Flora Salim}{UNSW}
    \icmlauthor{Xiaojie Yuan}{nk}
    \icmlauthor{Yanlong Wen}{nk}
    \icmlauthor{Ming Jin}{gu}
  \end{icmlauthorlist}

  \icmlaffiliation{nk}{Nankai University}
  \icmlaffiliation{tue}{Eindhoven University of Technology}
  \icmlaffiliation{ynu}{Yunnan University}
  \icmlaffiliation{dg}{DataDog}
  \icmlaffiliation{UNSW}{University of New South Wales}
  \icmlaffiliation{gu}{Griffith University}

  \icmlcorrespondingauthor{Yanlong Wen}{wenyl@nankai.edu.cn}
  \icmlcorrespondingauthor{Ming Jin}{mingjinedu@gmail.com}

  \icmlkeywords{Machine Learning, ICML}

  \vskip 0.3in
]



\printAffiliationsAndNotice{}  

\begin{abstract}
  Most time series foundation models are pretrained by directly predicting future observations, which often yields weakly structured latent representations that capture surface noise rather than coherent and predictable temporal dynamics. In this work, we introduce \modelname{}, a foundation model family that shifts pretraining from future value prediction to latent-space predictive learning. We train a causal Transformer to predict the evolution of latent representations, encouraging the emergence of structured and temporally coherent latent states. To ensure stable targets for latent-space learning, we design a lightweight aggregation branch to construct target representations. \modelname{} is optimized via a joint objective that integrates latent-space alignment, observational grounding to anchor representations to the input signal, and direct forecasting supervision. On the GIFT-Eval benchmark, \modelname{} mitigates structural fragmentation in the representation space and achieves state-of-the-art performance. These results demonstrate that constraining models to learn predictable latent dynamics is a principled step toward more robust and reliable time series foundation models. 

\end{abstract}

\input{Sections/Introduction}

\input{Sections/Related_Works}

\input{Sections/Methodology}

\input{Sections/Training_Data}

\input{Sections/Experiments}

\input{Sections/Conclusion}

\newpage

\input{Sections/Impact_Statement}

\bibliography{reference}
\bibliographystyle{icml2026}

\newpage

\input{Sections/Appendix}

\end{document}

%% file: Sections/Introduction.tex
\section{Introduction}
\label{sec:Introduction}

Pretrained foundation models have fundamentally changed the practice of time series forecasting. Instead of training a model for each time series (local models)~\citep{hyndman2008forecasting} or dataset (task-specific models)~\citep{lim2021temporal, challu2023nhits}, a foundation model can be trained once on large-scale time series data and then applied across different forecasting tasks~\citep{gruver2023large}. Foundation models not only streamline the forecasting workflow by eliminating repeated training, but also exhibit strong zero-shot capabilities, making them well suited to data-scarce regimes in which task-specific models often struggle to generalize~\citep{aksu2024gift}. In most existing time series foundation models, pretraining is performed via \emph{observation-space prediction}, where models are optimized to directly predict future values in the original signal space~\citep{ansari2025chronos2}.

\begin{figure}[t]
    \centering
    \includegraphics[width=0.45\textwidth]{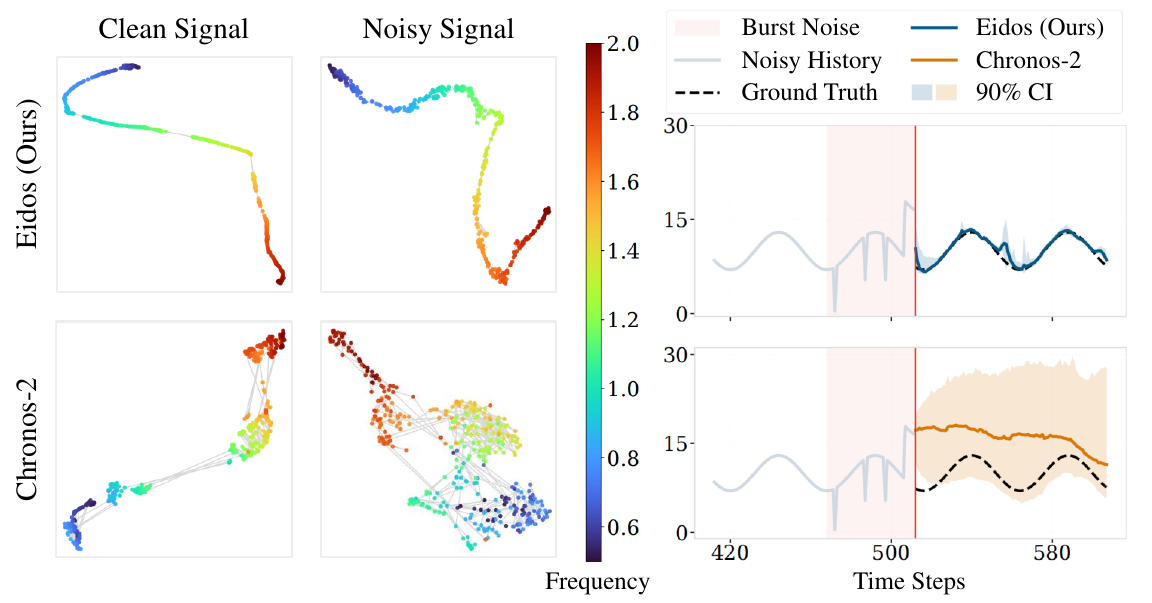}
    \caption{Latent-space (\modelname{}) versus observation-space (Chronos-2) prediction. (\textbf{Left}) \textit{Latent Representations}: UMAP visualization of the final-layer embeddings from each model under clean and noisy inputs, using time series with increasing frequencies. \modelname{} maintains a smooth and ordered manifold under noise, while Chronos-2 exhibits fragmented structure. (\textbf{Right}) \textit{Noise Robustness}: When forecasting under burst noise, the structured latent space of \modelname{} enables stable predictions, showing reduced sensitivity to surface-level noise observed in Chronos-2.}
    \label{fig:intro}
    \vspace{-10pt}
\end{figure}

Observation-space objectives focus primarily on matching future values and place little explicit constraint on how internal representations should be organized. As a result, the learned latent representations are often incidental byproducts of point-wise error minimization, without being explicitly encouraged to exhibit stable, structured, or temporally predictable dynamics. This decoupling implies that strong predictive performance does not necessarily correspond to well-organized latent states, and that representation quality is largely left to emerge implicitly during training~\citep{Wiliski2024ExploringRA}. In practice, this often leads to fragmented and temporally unstable latent spaces that are overly sensitive to noise and poorly suited for interpretation. As showcased in Figure \ref{fig:intro}, observation-space methods exhibit discontinuous manifolds that degrade under noisy inputs, making the model sensitive to surface-level irregularities rather than stable temporal structure.

This gap is particularly salient when compared to the evolution of foundation models in other modalities. In vision~\citep{Assran2025VJEPA2S}, language~\citep{Huang2025LLMJEPALL}, and speech~\citep{Ioannides2025JEPAAA}, pretraining paradigms have increasingly shifted from raw signal reconstruction toward latent-space predictive objectives~\citep{Assran2023SelfSupervisedLF}. Such a transition encourages models to move beyond surface-level noise and to capture structured transitions within latent representations. We argue that time series can often be viewed as noisy observations of an underlying dynamical system, making forecasting performance closely tied to how effectively a model internalizes these latent dynamics. A well-organized representation space is therefore not merely an auxiliary concern, but a central factor shaping the generalization, stability, and reliability of time series models, as qualitatively illustrated in Figure~\ref{fig:intro}.

We thus ask \textit{whether the training objective itself can be designed to explicitly constrain models to learn predictable latent dynamics, rather than only accurate observations}. We introduce \textbf{latent-space predictive learning}, which shifts the learning objective from future values to future representations. Concretely, models are trained to predict the temporal evolution of latent embeddings, encouraging the formation of structured and temporally coherent latent states.
Our formulation is deliberately minimalist: time series are mapped into a sequence of embeddings where a causal transformer is tasked solely with predicting the embeddings of subsequent time series. To ensure a stable target without collapse, we design a lightweight branch to construct the ground-truth embeddings, enforced by a stop-gradient bottleneck. 
Based on this paradigm, we introduce \modelname{}, a family of foundation models for time series forecasting that couple latent-space prediction with observation-space grounding through a joint objective. By avoiding heavy architectural priors in favor of predictive manifold alignment, our models achieve state-of-the-art performance while maintaining a concise architecture. Our contributions lie in these aspects:

\begin{itemize}
    \item We introduce latent-space predictive learning for time series foundation models, shifting the pretraining paradigm from observation space to latent space. 
    
    \item We present \modelname{}, a simple family of time series foundation models that unify representation learning and point forecasting through a joint prediction objective.
    
    \item Experimentally, \modelname{} mitigates noise overfitting and structural fragmentation, achieving top-tier performance and robustness on GIFT-Eval.
    
\end{itemize}

%% file: Sections/Related_Works.tex
 \section{Related Work}
\label{sec:Related_Works}



\textbf{Time Series Foundation Models} aim to acquire generalizable forecasting capabilities from large-scale time series corpora. Some approaches transfer sequence modeling abilities from large language models~\citep{Jin2023TimeLLMTS, gruver2023large} or pretrained vision models~\citep{Chen2024VisionTSVM, Shen2025VisionTSCT} to the time series domain. Others leverage the scalability of Transformer architecture~\citep{Yao2024TowardsNS} and directly train foundation models on extensive time series corpora. Within this family, encoder-only architectures include Chronos-2~\citep{ansari2025chronos2}, YingLong~\citep{wang2025output}, and Moirai v1~\citep{woo2024moirai}; decoder-only architectures include Moirai-MoE~\citep{liu2024moirai}, the TimesFM family~\citep{das2024decoder}, and Sundial~\citep{liu2025sundial}; and encoder–decoder models include Kairos~\citep{feng2025kairos}, Chronos and Chronos-Bolt~\citep{ansari2024chronos}, and FlowState~\citep{graf2025flowstate}. Beyond transformers, alternatives include TabPFN-TS~\citep{hoo2025tables}, based on a Prior-Data Fitted Network, and TiRex~\citep{auer2025tirex}, which builds on an xLSTM architecture. While current models prioritize architectural refinements, we argue that superior forecasting hinges not on heuristic increments, but on a fundamental reconstruction of the representation space.

\textbf{Time Series Representation Learning} is currently dominated by two paradigms: contrastive methods, which maximize instance-level similarity to capture global contexts~\citep{Li2024UniCLAU, Yue2021TS2VecTU}, and reconstruction methods, which prioritize data fidelity via encoder-decoder structures~\citep{Goswami2024MOMENTAF, Zerveas2020ATF}. In contrast, predictive learning~\citep{Rao1999PredictiveCI, Caron2021EmergingPI} remains underexplored in this field. Previous works such as Neural ODEs model latent dynamics through likelihood-driven objectives~\citep{Oh2024ComprehensiveRO}, where latent states serve as explanatory variables for generative modeling rather than prediction targets for representation learning. While Lat-PFN~\citep{Verdenius2024LaTPFNAJ} and TS-JEPA~\citep{Ennadir2025JointEG} recently introduced the Joint-Embedding Predictive Architecture (JEPA)~\citep{Assran2023SelfSupervisedLF}, they focused on building task-specific models rather than establishing a general pre-training paradigm. To extend the predictive learning pretraining paradigm to time series foundation models, we address two crucial challenges: eliminating the need for auxiliary encoders to streamline pre-training~\citep{Xu2025NextEmbeddingPM}, and preventing representation collapse to ensure embedding discriminability~\citep{Chen2020ASF}.



%% file: Sections/Methodology.tex
\section{Methodology}
\label{sec:Methodology}

In this section, we formalize the framework of latent-space predictive learning, followed by the architecture design and pretraining data of \modelname{}.

\subsection{Latent-Space Predictive Learning}

\begin{figure*}[t]
    \centering
    \begin{subfigure}{0.3\textwidth}
        \centering
        \includegraphics[width=\textwidth]{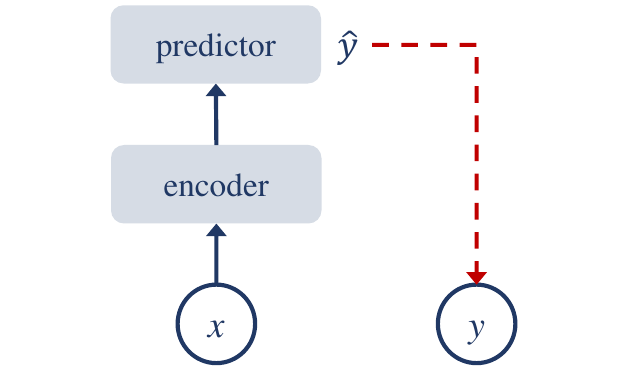}
        \caption{Next Token Prediction}
        \label{fig:3.1-1}
    \end{subfigure}
    \begin{subfigure}{0.3\textwidth}
        \centering
        \includegraphics[width=\textwidth]{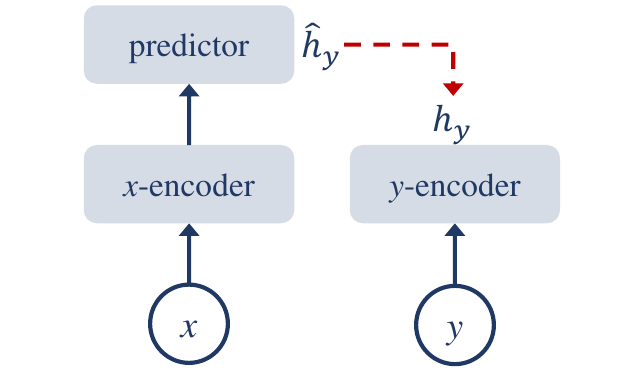}
        \caption{JEPA}
        \label{fig:3.1-2}
    \end{subfigure}
    \begin{subfigure}{0.3\textwidth}
        \centering
        \includegraphics[width=\textwidth]{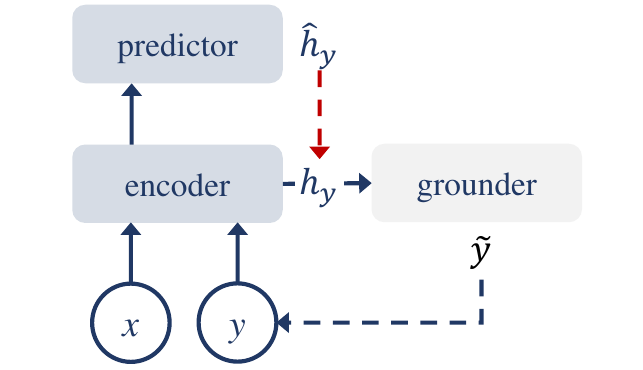}
        \caption{Ours}
        \label{fig:3.1-3}
    \end{subfigure}
    \label{fig:3.1}
    \caption{Comparison of predictive mechanisms. \textbf{(a) Next Token Prediction:} Direct prediction of future observations $y$ in the raw observation space. \textbf{(b) JEPA:} Latent-space alignment requiring a separate $y$-encoder to generate target representations $h_y$. \textbf{(c) Ours:} Latent-space prediction with a unified encoder, where target representations remain anchored to the raw observation $y$.}
\end{figure*}

\textbf{Next-Embedding Prediction.} Existing time series pretraining often collapses into raw signal prediction (Figure~\ref{fig:3.1-1}), which inadvertently captures stochastic noise rather than the underlying temporal structure. To move beyond raw signal noise, Joint-Embedding Predictive Architectures (JEPA) formulates prediction directly in the latent space, as shown in Figure~\ref{fig:3.1-2}. However, JEPA-style methods introduce significant training complexity; they rely on an auxiliary encoder to construct target embeddings from future observations.

As illustrated in Figure~\ref{fig:3.1-3}, to reduce the complexity introduced by the auxiliary encoder, we streamline JEPA into the simplest next-embedding prediction objective by removing the auxiliary encoder. Concretely, given a time series $\mathbf{x} \in \mathbb{R}^T$, we first map it into an embedding sequence $\mathbf{z} \in \mathbb{R}^{T \times d}$ using a point-wise embedding layer. We then learn an autoregressive predictor $f_\theta$ that predicts the latent representation of the future based on historical embeddings.
\begin{equation}
\widehat{\mathbf{h}}_{t+1} = f_\theta(\mathbf{z}_{\le t}),
\end{equation}
where $\widehat{\mathbf{h}}_{t+1}$ represents the predicted embedding associated with the next temporal position. 

\textbf{Latent Loss.} 
The predictor produces a single latent embedding for the next temporal position. For multi-step forecasting, supervision is defined over a future segment $\mathbf{z}_{t+1:t+l} \in \mathbb{R}^{l \times d}$. We therefore apply a lightweight aggregator $h_\phi$, implemented as a channel-wise convolution followed by an MLP, to map the future segment into a target embedding
$\mathbf{h}_{t+1} = h_\phi(\mathbf{z}_{t+1:t+l}) \in \mathbb{R}^{d}$. The alignment between the predicted and target embedding is optimized via negative cosine similarity, promoting a structured and directional latent space:
\begin{equation}
\mathcal{D}(\mathbf{h}, \widehat{\mathbf{h}}) = - \frac{1}{T-l} \sum_{t=1}^{T-l}\left( \frac{\mathbf{h}_{t+1}}{||\mathbf{h}_{t+1}||_2} \cdot \frac{\widehat{\mathbf{h}}_{t+1}}{||\widehat{\mathbf{h}}_{t+1}||_2} \right).
\end{equation}

A critical challenge in joint-embedding architectures is representation collapse. Following the asymmetric design principles of \citet{Chen2020ASF}, we apply a stop-gradient ($\operatorname{sg}$) operator to the target embedding. This ensures that the learning signal flows exclusively to the predictor, forcing it to adapt to the inherent structure of the future:
\begin{equation}
\mathcal{L}_{\text{latent}} = \mathcal{D}(\operatorname{sg}(\mathbf{h}), \widehat{\mathbf{h}}).
\end{equation}

\setlength{\algomargin}{0em} 
\SetAlFnt{\scriptsize} 
\begin{algorithm2e}[t]
    \PyCode{}\\
    \PyComment{f: predictor ($f_\theta$)} \\
    \PyComment{h: future aggregator ($h_\phi$)} \\
    \PyComment{g: grounding head ($g_\psi$)} \\
    \PyCode{} \\
    \PyKW{for} \PyCode{x} \PyKW{in} \PyCode{loader:} \\
    \Indp   
        \PyCode{z = embed(x)} \PyComment{input embeddings} \\ 
        \PyCode{h\_pred = f(z)} \PyComment{predicted latents} \\
        \PyCode{h\_target = h(z)} \PyComment{future latents} \\
        \PyCode{} \\
        \PyCode{latent\_loss = D(h\_pred, h\_target)} \PyComment{latent loss} \\
        \PyCode{gnd\_loss = Loss(g(h\_target))} \PyComment{grounding loss} \\
        \PyCode{loss = latent\_loss + gnd\_loss} \PyComment{total loss} \\
        \PyCode{} \\
        \PyCode{loss.backward()} \PyComment{back-propagate} \\
        \PyCode{update(f, h, g)} \PyComment{update parameters} \\
    \Indm 
    \PyCode{} \\
    \PyKW{def} \PyCode{D(pred, target):} \\
    \Indp
        \PyCode{target = target.detach()} \PyComment{stop gradient} \\
        \PyCode{} \\
        \PyCode{pred = normalize(pred, dim=-1)} \PyComment{l2-norm} \\
        \PyCode{target = normalize(target, dim=-1)} \PyComment{l2-norm} \\
        \PyCode{\PyKW{return} -(pred * target).\PyKW{sum}(dim=-1).mean()} \\
    \Indm
    \caption{Latent-Space Predictive Learning}
    \label{alg:lspl}
\end{algorithm2e}

\begin{figure}[t]
    \centering
    \includegraphics[width=0.4\textwidth]{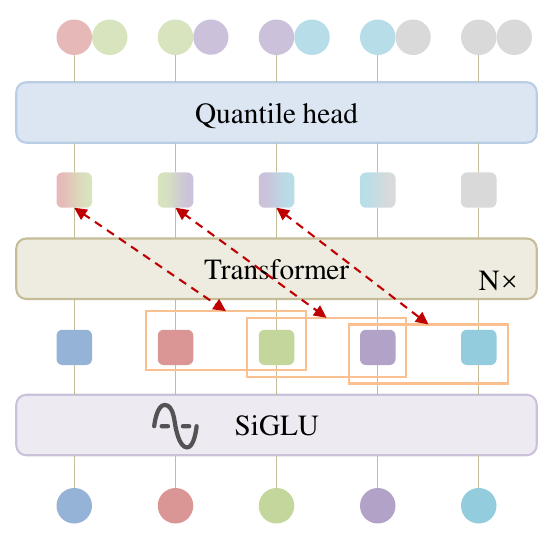}
    \label{fig:3.2}
    \caption{Architecture of \modelname{} with forecasting length $l=2$. The input time series is mapped to point-wise latent embeddings by SiGLU and processed by a causal Transformer to predict future latent states. A lightweight aggregator constructs target states from future segments, and predictions are aligned in the latent space. The quantile head operates on these latent representations to produce probabilistic forecasts over multiple future steps.}
\end{figure}

\textbf{Grounding Loss.}  To ensure that the aggregated representation $h_{t+1}$ remains grounded in the raw signals, we introduce an adjunct observation anchor. Specifically, a grounding head $g_\psi$ is tasked with restoring the original observation $\mathbf{x}_{t+1:t+l}$ from the latent target $\mathbf{h}_{t+1}$. This anchoring prevents the latent space from drifting into degenerate states:
\begin{equation}
\mathcal{L}_{\text{gnd}} = \frac{1}{T-l} \sum_{t=1}^{T-l} \mathcal{Q}\left(g_\psi(\mathbf{h}_{t+1}), \mathbf{x}_{t+1:t+l}\right),
\label{eq:4}
\end{equation}
where $\mathcal{Q}(\cdot,\cdot)$ is chosen as a quantile loss to align the grounding objective with downstream probabilistic forecasting. Together with the latent prediction objective, this grounding term provides a principled training signal that anchors latent evolution to observations, encouraging representations that remain informative for probabilistic forecasting. The latent-space predictive learning is summarized in Algorithm~\ref{alg:lspl}.

\subsection{\modelname{} Framework}

\subsubsection{Tokenization}

Given a univariate time series $\mathbf{x} \in \mathbb{R}^T$, we apply point-wise encoding to generate an embedding sequence $\mathbf{z} \in \mathbb{R}^{T \times d}$. Each scalar input $x_i$ is mapped to an embedding $\mathbf{z}_i$ via a sine-activated Gated Linear Unit (SiGLU):
\begin{equation}
\begin{aligned}
\mathbf{z}_i &= \text{GLU}(\sin(x_i \mathbf{W}_1 + b))\mathbf{W}_4, \\
\text{GLU}(\mathbf{h}) &= \text{Sigmoid}(\mathbf{h}\mathbf{W}_2) \odot \mathbf{h}\mathbf{W}_3,
\end{aligned}
\end{equation}
where $\mathbf{W}_{(\cdot)}$ and $b$ are learnable parameters, and $\odot$ denotes the element-wise product. The sine activation provides a set of bounded periodic basis functions, enabling a localized Fourier-like expansion of scalar inputs \cite{DBLP:conf/nips/SitzmannMBLW20,2022arXiv221201833N}, while the GLU adaptively gates these responses in a data-dependent manner. 

Point-wise tokenization maintains the temporal resolution of the input sequence, which simplifies latent-space prediction across time. Compared to patch-level aggregation, it avoids introducing an additional abstraction layer and keeps the prediction unit aligned with the raw signal.

\subsubsection{Model Backbone} 

The autoregressive predictor is implemented using a decoder-only Transformer with causal self-attention.
The model follows standard design choices for large-scale Transformers, including Pre-LN normalization~\cite{xiong2020layer} for stable optimization and rotary positional embeddings~\cite{su2024roformer} to encode relative temporal order. To support efficient long-sequence modeling~\cite{shoeybi2019megatron, rasley2020deepspeed}, we incorporate FlashAttention~\cite{dao2022flashattention} together with KV caching~\cite{pope2023efficiently}.

\subsubsection{Prediction Head and Training Objective}

The decoder output is mapped to the forecasting space through a residual prediction head that produces multi-step probabilistic forecasts.
Specifically, for each future time step $t$ and quantile level $q \in Q$, where $Q=\{0.1,\dots,0.9\}$, the model predicts a set of quantiles $\hat{y}_t^q$.
Model parameters are optimized using the standard quantile loss:
\begin{equation}
\small
\mathcal{L}_{\text{pred}}
=
\frac{1}{|Q| \cdot l}
\sum_{t, q}
\max \big[
q (y_t - \hat{y}_t^q),
(1-q)(\hat{y}_t^q - y_t)
\big],
\end{equation}
which encourages calibrated distributional forecasts across multiple horizons.

During pretraining, forecasting supervision is combined with latent-space predictive learning. The latent loss shapes the evolution of representations, while the forecasting loss anchors them to observed targets. For stability, the grounding head $g_\psi$ in Equation~\ref{eq:4} is implemented as a frozen copy of the forecasting head. Together, these objectives enable \modelname{} to learn structured temporal dynamics within a unified backbone, without additional architectural components.

%% file: Sections/Training_Data.tex
\subsection{Training Data}
\label{sec:Training_Data}


For a generalist pretrained model, the training data often plays a more decisive role than the model’s specific architecture. While recent studies~\citep{auer2025tirex, feng2025kairos} describe their data curation pipelines, no open dataset is yet available. To address this gap, we curated and released a time series pretraining corpus of over 170 billion time points to facilitate community progress. 
To strictly prevent data leakage, all real-world data is filtered to ensure zero overlap with downstream evaluation benchmarks. During pretraining, we sample from real and synthetic sources at a 1:1 ratio. For real-world data, we additionally apply TSMixup augmentation~\citep{ansari2024chronos} to improve robustness and diversity.

The corpus consists of three main components: (1) \textit{Chronos Training Data} ($0.8$ million time series):  We utilize a high-volume subset of the Chronos training repository~\citep{ansari2024chronos}, contributing approximately $83$ billion time points. (2) \textit{GiftEval Pre-training Data} ($2.5$ million time series): We incorporate a diverse collection from the GiftEval corpus~\citep{aksu2024gift}, adding $23$ billion time points. (3) \textit{Synthetic CauKer Data} ($15.7$ million time series): We generated $15.7$ million time series of length $4096$ using the CauKer method~\citep{xie2025cauker}, which samples from structural causal models based on Gaussian Processes to capture complex temporal dependencies.

%% file: Sections/Experiments.tex
\section{Experiments}
\label{sec:Experiments}

\begin{figure*}[t]
    \centering
    \includegraphics[width=\textwidth]{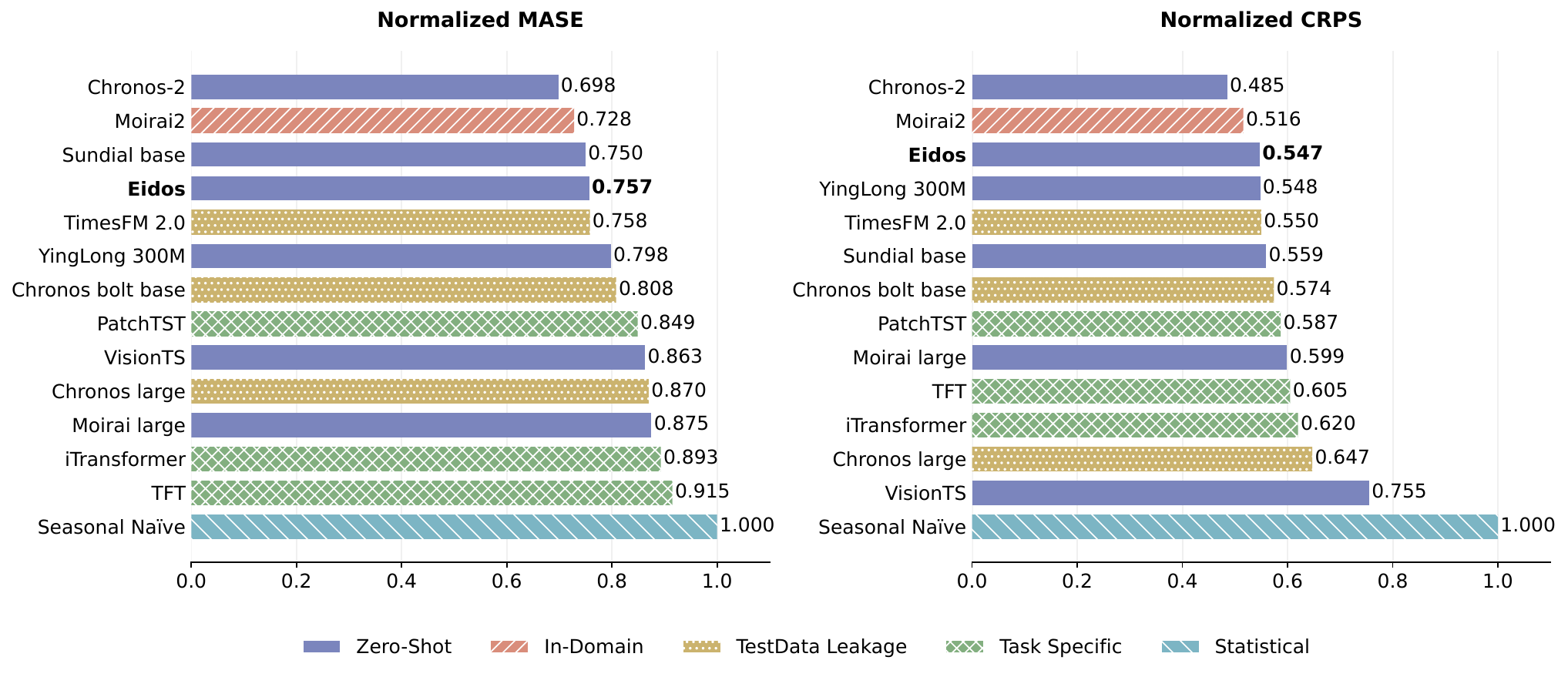}
    \caption{Performance on the GIFT-Eval benchmark across 97 evaluation configurations. Results are presented as Normalized MASE and Normalized CRPS where lower values indicate better accuracy. The legend distinguishes between different training paradigms. ``Zero-Shot'' models have no prior exposure to the benchmark data. ``In-Domain'' indicates that the benchmark training sets were included in the pre-training corpus, while ``TestData Leakage'' denotes models partially trained on the benchmark test sets. ``Task Specific'' and ``Statistical'' methods represent supervised and traditional baselines.}
    \label{fig:gift_eval}
\end{figure*}

This section presents a systematic empirical evaluation of \modelname{}, focusing on the benefits of latent-space predictive learning.
We first report zero-shot forecasting performance on standardized benchmarks (Section~\ref{sec:zs-results}).
We then analyze robustness under input perturbations (Section~\ref{sec:noise_robustness_analysis}) and examine the structure and controllability of the learned representations through probing and latent interventions (Section~\ref{sec:representation_exp}).
Finally, we conduct ablation and scaling studies to clarify the role of key design choices and model capacity (Sections~\ref{sec:ablation-results} and~\ref{sec:scalability}), followed by a brief discussion in Section~\ref{sec:Discussions}.
Additional details and extended results are provided in Appendix~\ref{app:experiment_details} and Appendix~\ref{app:extended_results}.


\textbf{\modelname{} Training} 
We train \modelname{} for 100,000 steps using a global batch size of 256. The optimization employs AdamW with $\beta_1=0.9$, $\beta_2=0.95$, and a weight decay of $1 \times 10^{-2}$. The learning rate is set to $1 \times 10^{-3}$ and follows a schedule with 10,000 linear warmup steps before transitioning to cosine annealing. For computational efficiency, we implement bfloat16 mixed precision. We train \modelname{} with context length of 4096 and forecasting length of 64. During inference, we use the 0.5 quantile for point forecasting and generate predictions autoregressively.

\textbf{Evaluation Benchmarks} We primarily evaluate \modelname{} on the GIFT-Eval benchmark~\citep{aksu2024gift}. 
GIFT-Eval comprises $23$ datasets evaluated under multiple settings and spans short-, medium-, and long-term forecasting horizons across different sampling frequencies, resulting in a total of $97$ evaluation configurations. In addition, we provide results on the recently introduced fev-bench~\citep{shchur2025fev} in Appendix~\ref{app:fev-bench}.

\textbf{Evaluation Protocol} The evaluation follows the respective benchmark protocols using mean absolute scaled error (MASE) for point forecast performance and the continuous ranked probability score (CRPS) for probabilistic forecast performance. 
Aggregated performance is computed by normalizing each evaluation setting's score by that of a seasonal na\"ive baseline, followed by the geometric mean across evaluation settings. 
Additionally, the average rank across evaluation settings is reported to ensure robustness against outlier performance.


\subsection{Benchmark Results}\label{sec:zs-results} 


The zero-shot performance of \modelname{} on GIFT-Eval is summarized in Figure~\ref{fig:gift_eval}. Overall, \modelname{} demonstrates competitive results, positioning itself as a robust mid-to-high tier forecaster among current foundation models. In point forecasting, \modelname{} achieves a normalized MASE of 0.757, performing on par with Sundial base and delivering a 10.8\% reduction in error compared to the task-specific PatchTST. This highlights the strong generalization ability of our zero-shot approach over supervised models trained on in-distribution data.

In terms of probabilistic forecasting, \modelname{} achieves a normalized CRPS of 0.547, securing a top-tier position among zero-shot models. In this metric, our model outperforms several established baselines, including TimesFM 2.0 and YingLong 300M. While a gap remains relative to the current state-of-the-art Chronos-2, \modelname{} maintains competitive performance across both point and probabilistic metrics with a substantially simpler design. A detailed efficiency comparison is reported in Appendix~\ref{app:model_efficiency}.

\begin{takeaway}
\textbf{Takeaway:} \modelname{} achieves competitive zero-shot forecasting performance using a comparatively lightweight architecture.
\end{takeaway}



\subsection{Noise Robustness Analysis}
\label{sec:noise_robustness_analysis}

\begin{figure}[t]
    \centering
    \begin{subfigure}{0.235\textwidth}
        \centering
        \includegraphics[width=\textwidth]{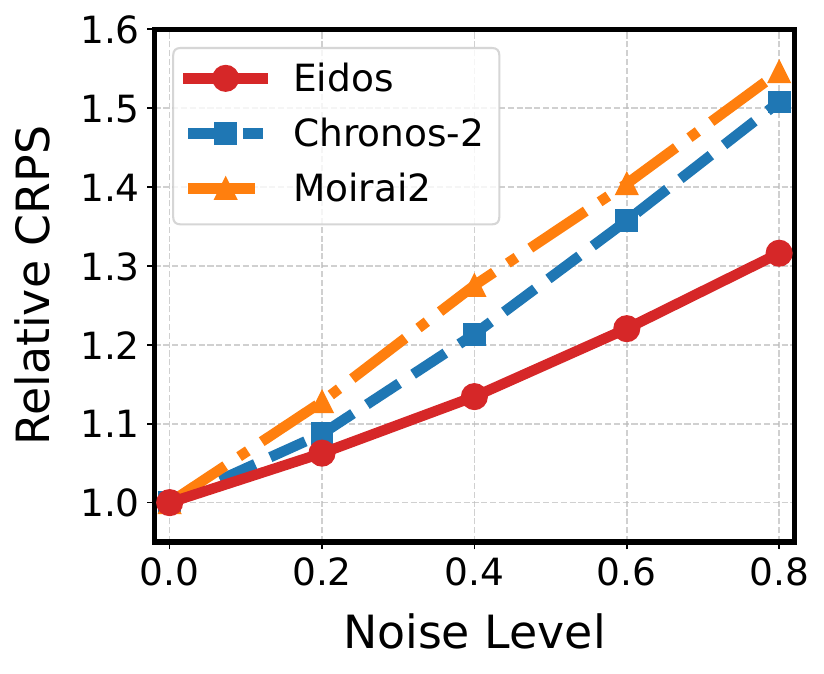}
        \caption{Gaussian Noise}
        \label{fig:gaussian_noise}
    \end{subfigure}
    \begin{subfigure}{0.235\textwidth}
        \centering
        \includegraphics[width=\textwidth]{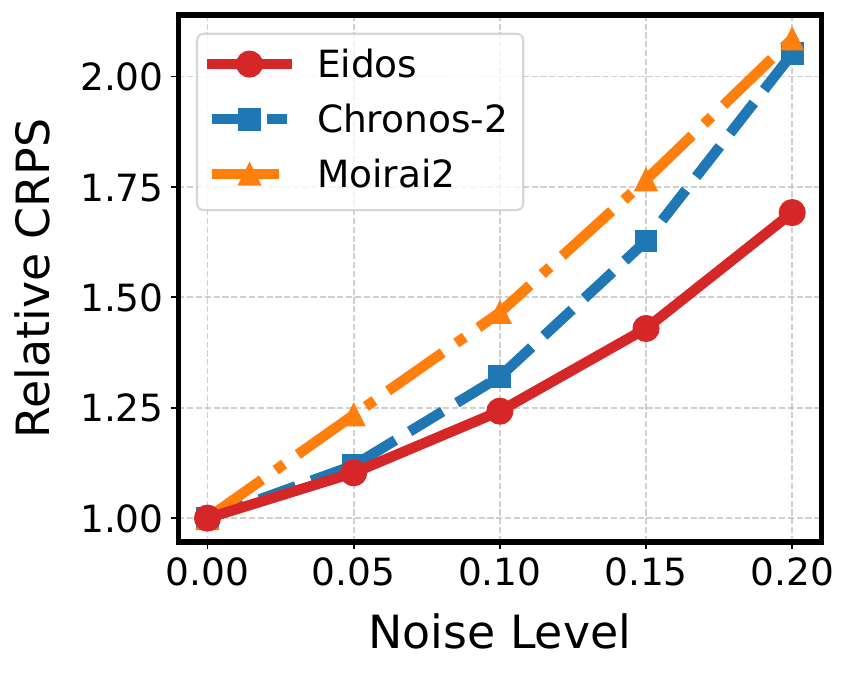}
        \caption{Impulse Noise}
        \label{fig:impulse_noise}
    \end{subfigure}
    \caption{Noise Robustness Analysis. Relative CRPS degradation under increasing intensities of Gaussian and impulse noise. All results are normalized to the clean baseline performance for each model.}
    \label{fig:noise_degradation}
\end{figure}

We evaluate noise robustness on the GIFT-Eval benchmark by injecting two types of perturbations into the input context, including additive Gaussian noise with intensities from 0.0 to 0.8 scaled by series standard deviation and impulse noise with intensities from 0.0 to 0.2. Figure~\ref{fig:noise_degradation} reports the relative degradation in predictive performance measured by Relative CRPS, normalized to the clean baseline.

As noise levels increase, the performance of observation-space baselines Chronos-2 and Moirai2 degrades rapidly across both regimes. In the Gaussian noise case shown in Figure~\ref{fig:gaussian_noise}, baseline errors increase by over 50 percent at the highest intensity. \modelname{} limits this degradation to approximately 1.31. A similar trend is observed under impulse noise in Figure~\ref{fig:impulse_noise} where baseline degradation exceeds 2.0 while \modelname{} remains significantly more stable at approximately 1.7. These results suggest that our latent-space predictive objective acts as an effective bottleneck, suppressing non-predictive stochastic noise while preserving stable temporal dynamics essential for reliable forecasting. 
\begin{takeaway}
    \textbf{Takeaway:} Latent-space predictive learning yields forecasts that are markedly more stable under input noise than observation-space objectives.
\end{takeaway}

\subsection{Representational Analysis and Intervention}
\label{sec:representation_exp}

\begin{figure}[t]
    \centering
    \begin{subfigure}{0.235\textwidth}
        \centering
        \includegraphics[width=\textwidth]{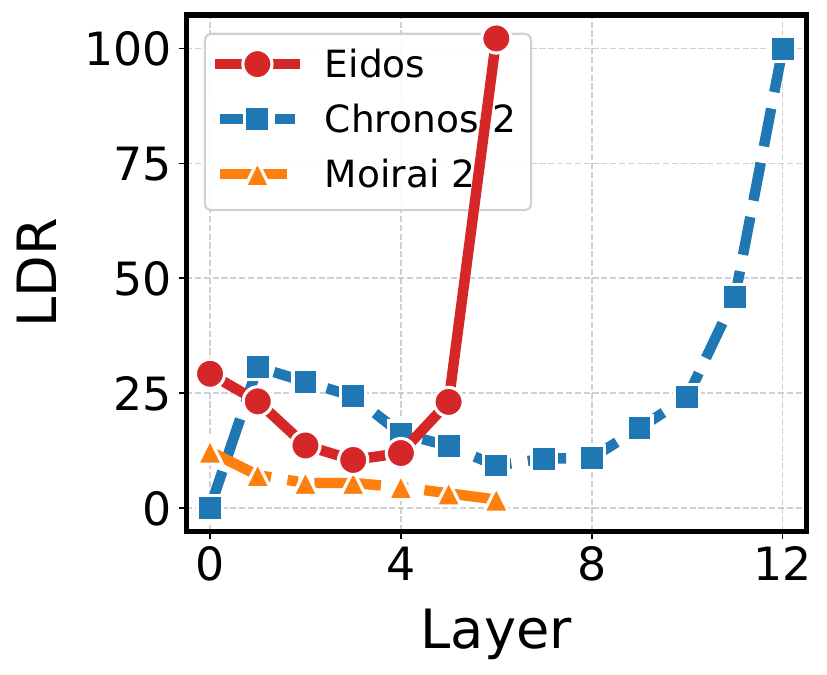}
        \caption{Trend LDR}
        \label{fig:trend_ldr}
    \end{subfigure}
    \begin{subfigure}{0.235\textwidth}
        \centering
        \includegraphics[width=\textwidth]{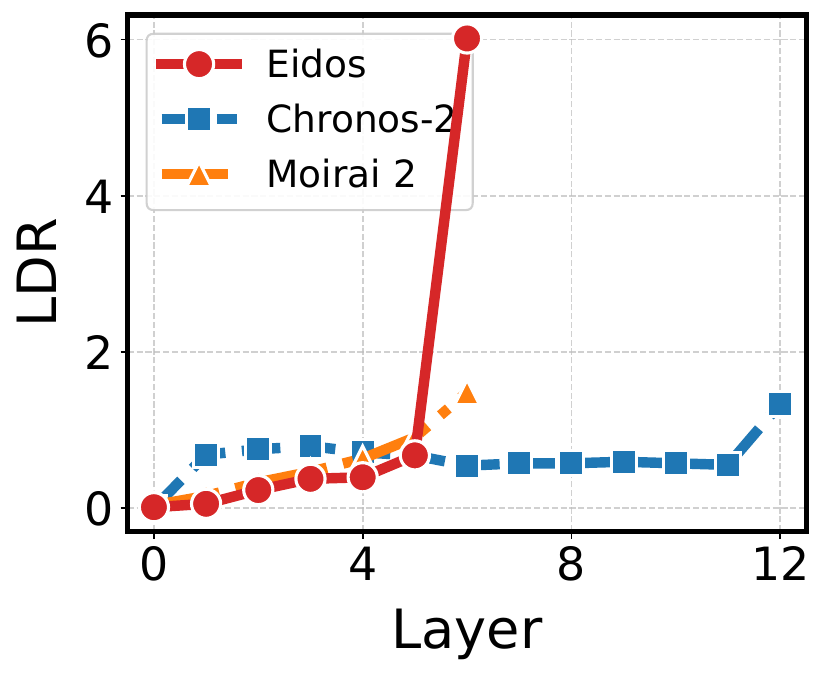}
        \caption{Periodicity LDR}
        \label{fig:period_ldr}
    \end{subfigure}
    \caption{Layer wise Feature Probing. Evolution of the Linear Discriminant Ratio (LDR) for trend and periodicity attributes across hidden layers. The metric measures the linear separability of fundamental temporal features.}
    \label{fig:probe}
\end{figure}

\paragraph{Layer-wise Feature Probing.} To investigate how models organize internal information, we analyze the linear separability of representations using the Linear Discriminant Ratio (LDR). Figure~\ref{fig:probe} compares the LDR trajectories for trend and periodicity features across layers, revealing a consistent advantage for \modelname{}.

In the trend detection task shown in Figure~\ref{fig:trend_ldr}, \modelname{} exhibits a pronounced increase in LDR at its final layer, achieving strong separability despite utilizing a compact 6-layer backbone. In comparison, the deeper Chronos-2 requires twelve layers to reach comparable separation, while Moirai 2 remains largely inseparable. A similar pattern is observed for periodicity in Figure~\ref{fig:period_ldr}. Overall, these results indicate that latent-space predictive learning encourages the encoding of fundamental temporal properties into representations that are more linearly organized and easier to separate. 
\begin{takeaway}
    \textbf{Takeaway:} Predictive supervision in latent space leads to more structured and linearly separable representations, even with a shallow backbone.
\end{takeaway}

\begin{figure*}[t]
    \centering
    \begin{subfigure}{0.235\textwidth}
        \centering
        \includegraphics[width=\textwidth]{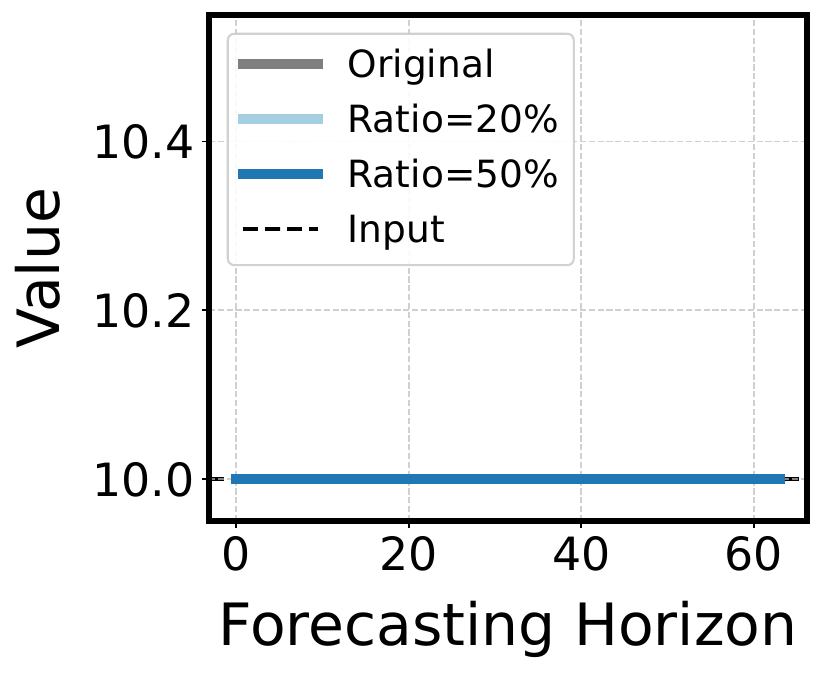}
        \caption{Trend, Chronos-2}
        \label{fig:trend_chronos2}
    \end{subfigure}
    \begin{subfigure}{0.235\textwidth}
        \centering
        \includegraphics[width=\textwidth]{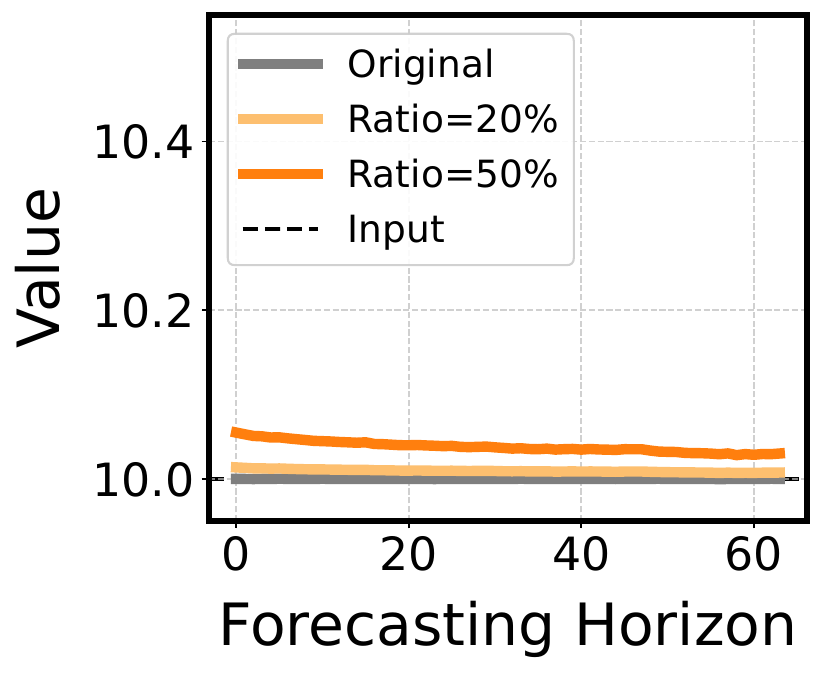}
        \caption{Trend, Moirai2}
        \label{fig:trend_moirai2}
    \end{subfigure}
    \begin{subfigure}{0.235\textwidth}
        \centering
        \includegraphics[width=\textwidth]{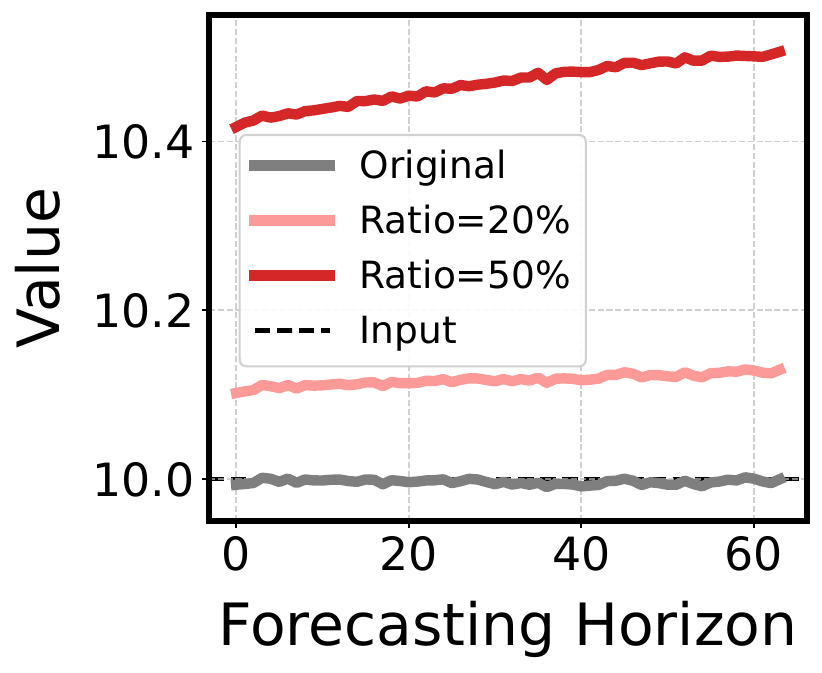}
        \caption{Trend, \modelname{}}
        \label{fig:trend_ours}
    \end{subfigure}
    \begin{subfigure}{0.235\textwidth}
        \centering
        \includegraphics[width=\textwidth]{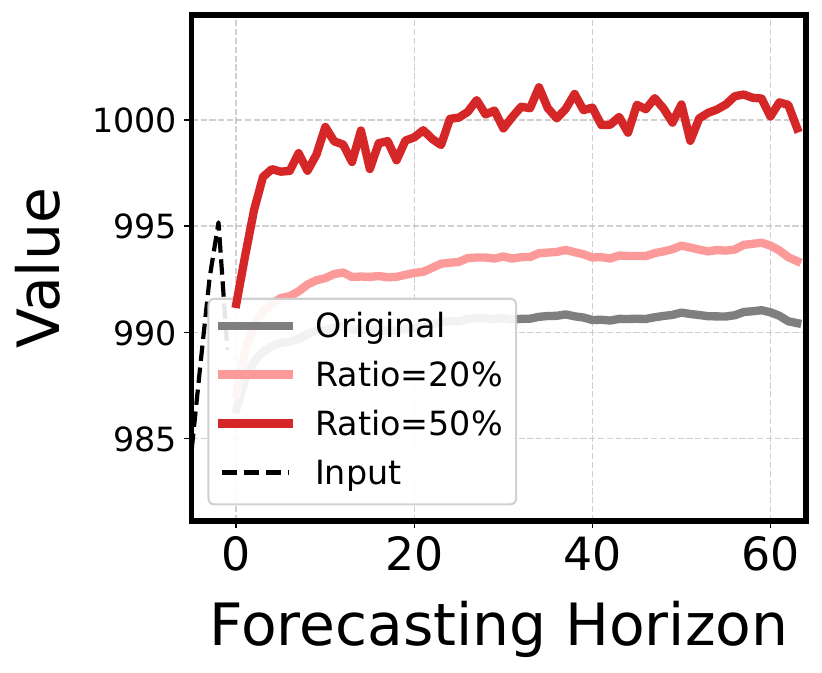}
        \caption{Trend, \modelname{}, Jena Weather}
        \label{fig:trend_ours}
    \end{subfigure}
    \begin{subfigure}{0.235\textwidth}
        \centering
        \includegraphics[width=\textwidth]{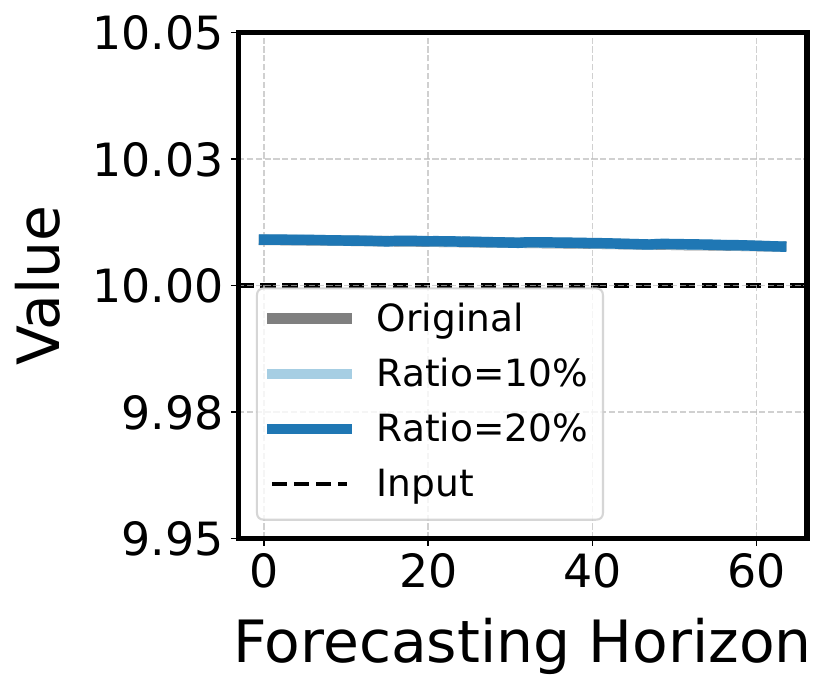}
        \caption{Periodicity, Chronos-2}
        \label{fig:period_ours}
    \end{subfigure}
    \begin{subfigure}{0.235\textwidth}
        \centering
        \includegraphics[width=\textwidth]{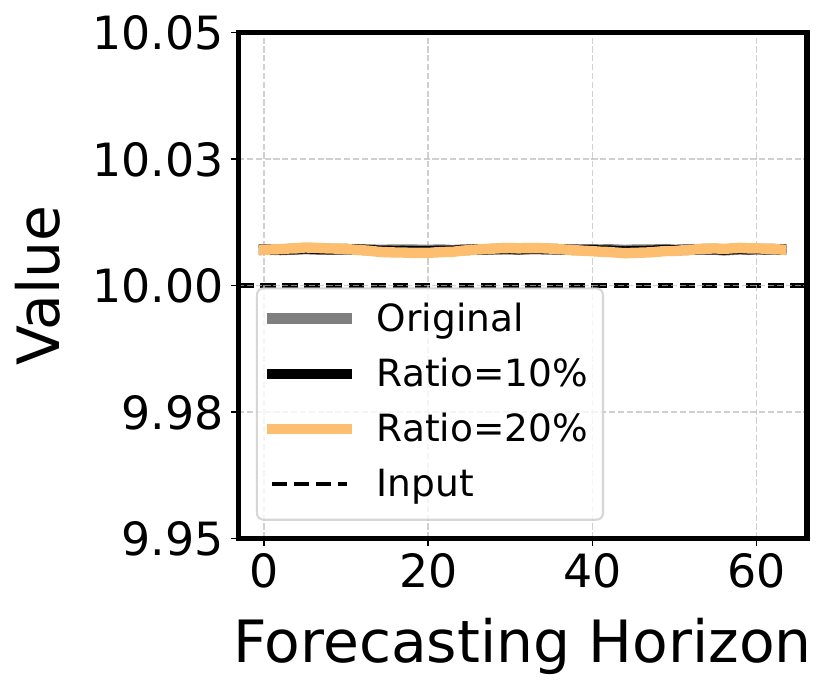}
        \caption{Periodicity, Moirai2}
        \label{fig:period_ours}
    \end{subfigure}
    \begin{subfigure}{0.235\textwidth}
        \centering
        \includegraphics[width=\textwidth]{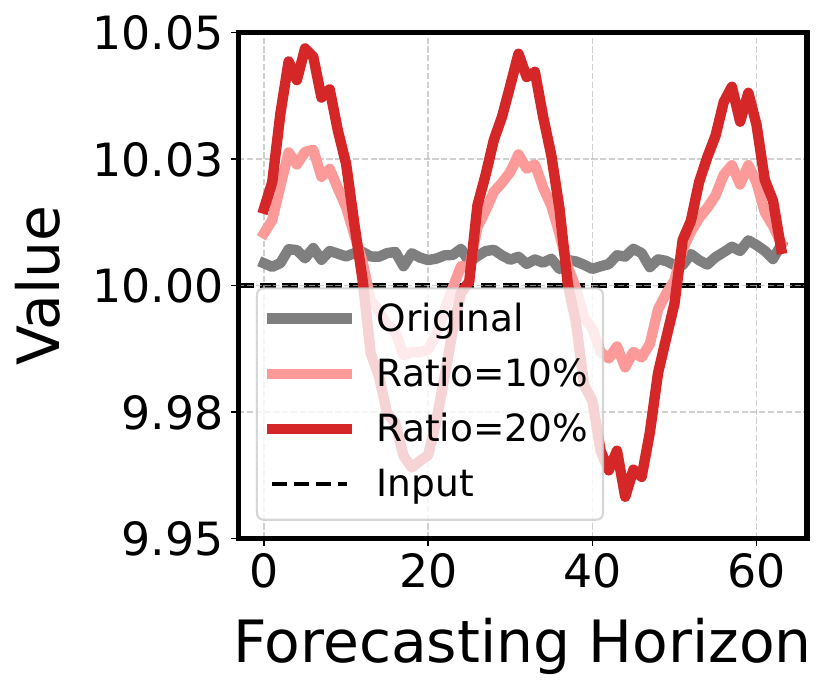}
        \caption{Periodicity, \modelname{}}
        \label{fig:period_ours}
    \end{subfigure}
    \begin{subfigure}{0.235\textwidth}
        \centering
        \includegraphics[width=\textwidth]{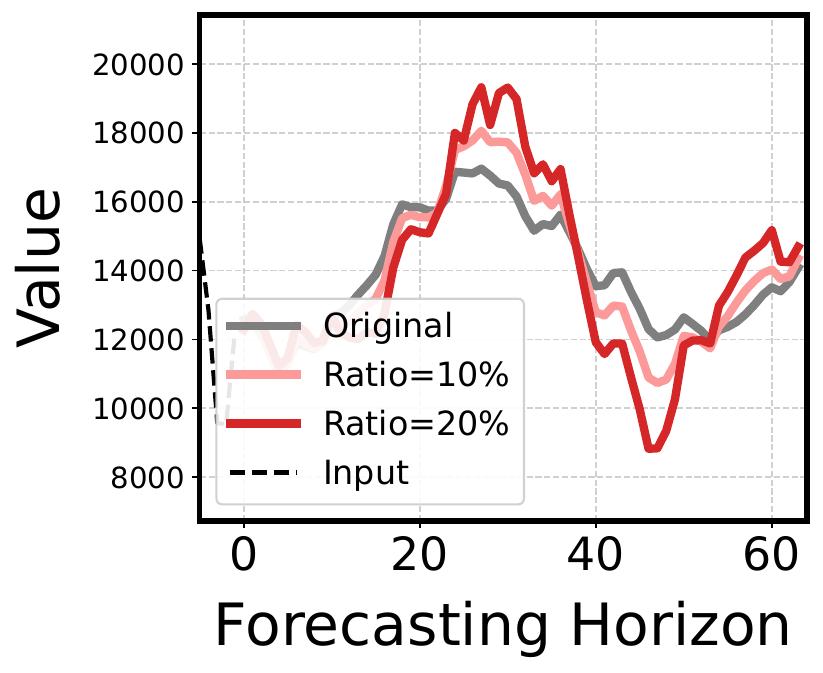}
        \caption{Periodicity, \modelname{}, Solar}
        \label{fig:period_ours}
    \end{subfigure}
    \caption{Latent space steering. Trend and periodicity directions identified through linear probing are added to the final-layer hidden states with varying magnitudes. Original predictions are compared against modified outputs to show the results of latent space manipulation prior to the forecasting head. Results include baseline comparisons on synthetic signals, with \modelname{} extending to real-world datasets.}
    \label{fig:steer}
\end{figure*}

\paragraph{Latent Steering and Editing.} To examine whether the learned representations play a functional role in generation, we conduct latent steering experiments by injecting characteristic directions obtained from the probing tasks into the hidden states of a neutral input. Figure~\ref{fig:steer} reports the resulting forecasts under trend and periodicity interventions. We further show that the same latent interventions remain effective on real-world time series.

For \modelname{}, these interventions produce clear and systematic effects. Steering along the trend direction converts an otherwise flat prediction into a stable upward trajectory, while periodicity steering induces coherent oscillatory patterns. We observe that the same interventions induce meaningful and interpretable changes on real-world time series. Moreover, the strength of the output response varies smoothly with the injection magnitude, suggesting that semantic directions in the representation space are continuously and predictably reflected in the generated time series.

By comparison, the baseline models exhibit little to no response. Chronos-2 and Moirai2 largely retain flat predictions under the same interventions, indicating a weak coupling between their latent features and explicit temporal dynamics. The contrasting behavior of \modelname{} suggests that its representations encode fundamental temporal attributes in a form that is directly usable for controlled editing, rather than remaining implicit or entangled.
\begin{takeaway}
    \textbf{Takeaway:} The learned latent representations support direct and predictable control over forecasting behavior.
\end{takeaway}

\subsection{Ablation Study}\label{sec:ablation-results}

\begin{table*}[htbp]
  \centering
  \caption{Ablation study on key components of \modelname{} across short, medium, and long term horizons on GIFT-Eval. Results are reported for the base model (v0), the addition of SiGLU encoding (v1), latent loss (v2), and grounding loss (\modelname{}). Bold indicates best results and the final row quantifies the total error reduction relative to v1.}
  \footnotesize 
  \definecolor{gain}{HTML}{31a354} 
  \definecolor{highlight}{HTML}{f0f7f0} 
    \begin{tabular}{llllrrrrrrrr}
    \toprule
    \multicolumn{1}{c}{\multirow{2}{*}{Variant}} & \multicolumn{1}{c}{\multirow{2}{*}{SiGLU}} & \multicolumn{1}{c}{\multirow{2}{*}{$\mathcal{L}_{\text{latent}}$}} & \multicolumn{1}{c}{\multirow{2}{*}{$\mathcal{L}_{\text{gnd}}$}} & \multicolumn{2}{c}{Short} & \multicolumn{2}{c}{Medium} & \multicolumn{2}{c}{Long} & \multicolumn{2}{c}{Overall} \\
    \cmidrule(lr){5-6} \cmidrule(lr){7-8} \cmidrule(lr){9-10} \cmidrule(lr){11-12}
       &    &    &    & MASE  & CRPS  & MASE  & CRPS  & MASE  & CRPS  & MASE  & CRPS \\
    \midrule
    v0 & ---  & ---  & ---  & 0.7359 & 0.5489 & 0.8218 & 0.5844 & 0.8448 & 0.5895 & 0.7766 & 0.5651 \\
    v1 & \checkmark  & ---  & ---  & 0.7290 & 0.5443 & 0.8032 & 0.5778 & 0.8408 & 0.5832 & 0.7678 & 0.5597 \\
    v2 & \checkmark  & \checkmark  & ---  & 0.7314 & 0.5461 & 0.8058 & 0.5732 & 0.8307 & 0.5703 & 0.7678 & 0.5570 \\
    \modelname{} & \checkmark  & \checkmark  & \checkmark  & \textbf{0.7238} & \textbf{0.5395} & \textbf{0.7859} & \textbf{0.5583} & \textbf{0.8198} & \textbf{0.5556} & \textbf{0.7569} & \textbf{0.5470} \\
    \midrule
    \multicolumn{4}{l}{\textit{Rel. Gain vs. v1}} & {\color{gain}0.7\%$\downarrow$} & {\color{gain}0.9\%$\downarrow$} & \cellcolor{highlight}{\color{gain}2.2\%$\downarrow$} & \cellcolor{highlight}{\color{gain}3.4\%$\downarrow$} & \cellcolor{highlight}{\color{gain}2.5\%$\downarrow$} & \cellcolor{highlight}{\color{gain}4.7\%$\downarrow$} & {\color{gain}1.4\%$\downarrow$} & {\color{gain}2.3\%$\downarrow$} \\
    \bottomrule
    \end{tabular}
  \label{tab:ablation}
\end{table*}

We conduct an ablation study to examine the contribution of each component in \modelname{}. Table~\ref{tab:ablation} reports results on GIFT-Eval across short-, medium-, and long-term horizons. Comparing v0 and v1, introducing SiGLU encoding consistently improves both MASE and CRPS across all horizons. This indicates that sine-based point-wise tokenization provides a more effective representation of temporal signals than standard embeddings.

Variant v2 further adds the latent-space prediction objective. Its effect varies across prediction horizons, with limited changes at short and medium ranges and clearer improvements at long horizons. This indicates that latent-space prediction alone does not always improve performance when it is weakly coupled to the original signal.

The full \modelname{} achieves the best performance by incorporating the grounding loss $\mathcal{L}_{\text{gnd}}$. The grounding loss provides direct supervision from the original signal, ensuring that latent predictions remain tied to numerical forecasting objectives. Compared to v1, the full model shows more pronounced improvements at medium and long horizons, demonstrating the advantage of predicting in the latent space rather than relying on observation-space supervision. 

Overall, these results show that the effectiveness of \modelname{} arises from the joint design of expressive tokenization, latent prediction, and observational grounding, rather than from any single component in isolation.

\subsection{Scaling Experiments} \label{sec:scalability}

\begin{table}[htbp]
  \centering
  \caption{Performance of different \modelname{} sizes.}
  \footnotesize 
    \begin{tabular}{lccc}
    \toprule
    \textbf{Model} & \multicolumn{1}{l}{\textbf{Params (M)}} & \multicolumn{1}{l}{\textbf{MASE}} & \multicolumn{1}{l}{\textbf{CRPS}} \\
    \midrule
    \modelname{} Small & 12.7 & 0.7569 & 0.5470 \\
    \modelname{} Base & 23.3 & 0.7650 & 0.5591 \\
    \modelname{} Large & 91.7 & 0.7523 & 0.5514 \\
    \bottomrule
    \end{tabular}%
  \label{tab:scalibility}%
\end{table}%

We study the effect of model capacity by training three variants of \modelname{} with parameter counts ranging from 12.7M to 91.7M. Table~\ref{tab:scalibility} reports the corresponding results.
We do not observe a clear monotonic relationship between model size and performance. While the largest variant achieves the lowest MASE, the smallest model attains the best CRPS, and overall differences across scales remain modest.

These results indicate that increasing parameter count alone does not consistently improve forecasting performance, particularly for probabilistic accuracy. Notably, the smallest variant already performs competitively, indicating that the proposed latent-space predictive objective can be effectively realized with relatively limited capacity. This highlights the parameter efficiency of \modelname{}, where modeling temporal evolution in latent space reduces the reliance on large overparameterized models.

Similar behavior has been reported in recent studies on time series foundation models, where scaling effects are often less predictable than in language modeling \cite{liu2025moirai, aksu2024gift}. In this setting, further improvements are therefore more likely to depend on advances in data, learning objectives, or architectural design, rather than on increasing model size alone.

\subsection{Discussions}
\label{sec:Discussions}

\begin{figure}[t]
    \centering
        \begin{subfigure}{0.235\textwidth}
        \centering
        \includegraphics[width=\textwidth]{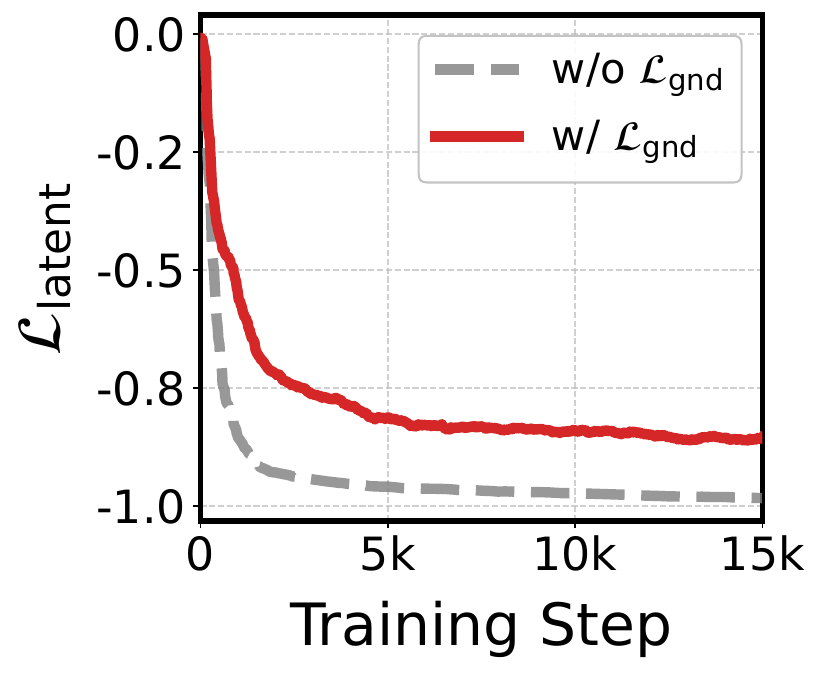}
        \caption{Impact of $\mathcal{L}_{\text{gnd}}$}
        \label{fig:discuss_recloss}
    \end{subfigure}
    \begin{subfigure}{0.235\textwidth}
        \centering
        \includegraphics[width=\textwidth]{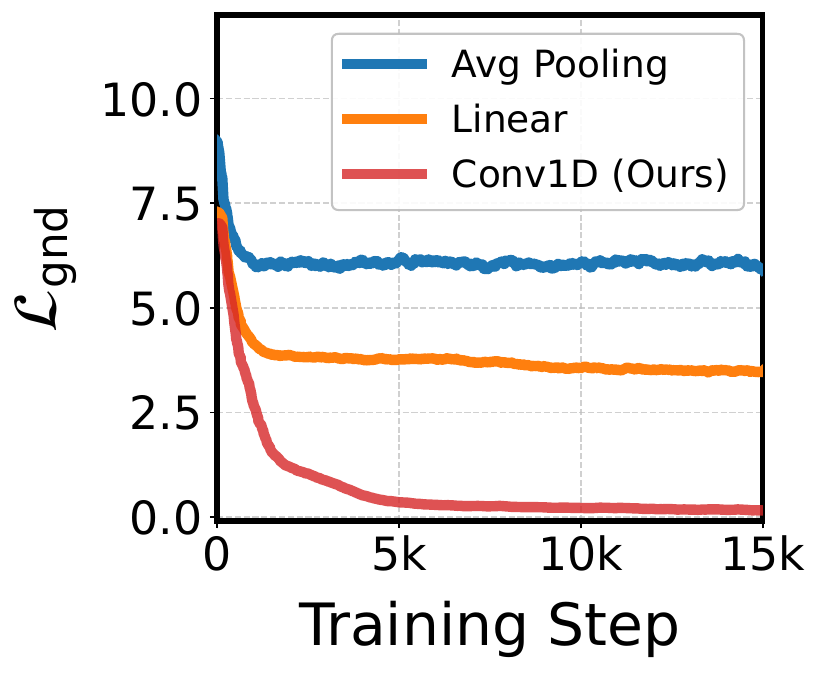}
        \caption{Aggregator comparison}
        \label{fig:discuss_aggregator}
    \end{subfigure}
    \caption{Training loss trajectories. Latent loss convergence with and without the grounding constraint and grounding loss for Conv1D, Average Pooling, and Linear aggregator designs.}
    \label{fig:ablation}
\end{figure}

\paragraph{Prevention of Representation Collapse.} 
Figure~\ref{fig:discuss_recloss} visualizes the latent loss trajectories. The variant without grounding loss (w/o $\mathcal{L}_{\text{gnd}}$, v2 in Section~\ref{sec:ablation-results}) converges to a substantially lower value. Although this may appear favorable in terms of objective minimization, it is indicative of representation collapse. In the absence of an observational anchor, the predictor and aggregator can satisfy the alignment objective through near-constant latent solutions. By contrast, the full model (w/ $\mathcal{L}_{\text{gnd}}$) maintains a higher loss plateau. As reflected by its consistently better forecasting performance in Table~\ref{tab:ablation}, the grounding constraint discourages such degenerate behavior and encourages the preservation of informative latent structure.

\paragraph{Channel-Specific Aggregation.} 
We implement the aggregator as a depthwise convolution to learn channel-specific temporal kernels. Figure~\ref{fig:discuss_aggregator} contrasts this design with Avg Pooling and Linear baselines. Avg Pooling yields the highest loss, as it removes temporal ordering through uniform averaging. The Linear baseline improves upon this but remains constrained by sharing a single temporal projection across all channels. The stronger convergence of the depthwise design suggests that different latent dimensions follow distinct temporal dynamics, motivating the use of independent aggregation mechanisms rather than a unified temporal summary.

%% file: Sections/Conclusion.tex
\section{Conclusion}
\label{sec:Conclusion}

In this work, we presented \modelname{}, a time series foundation model that reframes pretraining from observation-space prediction to latent-space predictive learning. By explicitly modeling the evolution of representations, \modelname{} captures coherent temporal dynamics while reducing sensitivity to surface-level noise. The proposed joint objective integrates latent prediction with grounding and forecasting, preventing representation collapse and supporting accurate predictions. Empirical results on GIFT-Eval demonstrate state-of-the-art performance, along with improved robustness, feature organization, and controllability in latent space. These findings highlight the importance of learning predictable latent dynamics for reliable time series modeling, and point toward latent-space predictive learning as a promising direction for future foundation models.

\paragraph{Limitations \& Future Work} We focus on univariate forecasting, following recent large-scale pretraining paradigms and benchmarks \cite{liu2025moirai, liu2025sundial}. Extending the latent-space predictive framework to multivariate settings or covariates remains a natural direction \cite{ansari2025chronos2}. Moreover, model design and hyperparameters prioritize architectural simplicity over exhaustive tuning or data curation, leaving room for further optimization in large-scale or production-oriented settings. Beyond forecasting, explicitly modeling temporal evolution in latent space provides a structured basis for representation analysis and latent reasoning over long horizons. This latent formulation also offers a natural interface for incorporating contextual or multimodal signals, such as text or structured metadata, enabling richer conditioning and more interpretable interactions with temporal models.

%% file: Sections/Impact_Statement.tex
\section*{Impact Statement}
\label{sec:impact_statement}

Our work focuses on improving the reliability and representational quality of time series foundation models. By emphasizing latent-space dynamics over observation-space reconstruction, \modelname{} reduces model sensitivity to input noise, a common failure mode in real-world forecasting. The research is methodological and utilizes publicly available, non-sensitive data. We believe the move toward more structured latent spaces represents a positive step for the stability of automated decision-making systems. There are no specific ethical concerns or negative societal impacts associated with this work that are distinct from the general challenges of AI-based predictive modeling.

%% file: Sections/Appendix.tex
\appendix
\onecolumn
\section{Experiment Details}
\label{app:experiment_details}

\subsection{Experimental Settings}
This section provides a description of the architectural configurations, optimization hyperparameters, and evaluation protocols used in our study. 

\subsubsection{Model Architecture and Hyperparameters}
The primary \modelname{} model used in the main experiments (referred to as \textit{\modelname{} small}) is based on a decoder-only Transformer backbone. The specific architectural and training parameters are detailed below.

\paragraph{Architecture details.} 
The input sequence is first processed by a SiGLU-based point-wise tokenizer. The Transformer backbone consists of 6 decoder layers with a hidden dimension of 384 and an intermediate feed-forward dimension of 1024. We utilize Multi-Head Attention (MHA) with 12 attention heads, resulting in a head dimension of 32. To encode relative temporal order, we employ Rotary Positional Embeddings (RoPE) with a base frequency $\theta = 10,000$. For normalization, we use LayerNorm with $\epsilon = 1e-6$ in a Pre-LN configuration. The SiGLU activation is used in the tokenizer, while the Transformer blocks utilize the SiLU activation function within the gated MLP layers. The probabilistic forecasting head produces 9 quantiles $Q = \{0.1, 0.2, 0.3, 0.4, 0.5, 0.6, 0.7, 0.8, 0.9\}$.

\paragraph{Latent Predictive Learning.}
The latent aggregator is implemented as a depthwise 1D convolution with a kernel size matching the horizon length $l=64$, distilling temporal segments into a single target embedding. The joint objective is optimized with weights $\lambda_{latent} = 0.1$ and $\lambda_{gnd} = 0.1$, while the primary forecasting loss ($\mathcal{L}_{pred}$) has a weight of $1.0$. We apply a stop-gradient operator to the target branch of the latent-space alignment to prevent representation collapse.

\paragraph{Training Setup.}
\modelname{} is trained for 100,000 iterations. We use the AdamW optimizer with $\beta_1 = 0.9, \beta_2 = 0.95$ and a weight decay of $0.01$. The learning rate is initialized at $1 \times 10^{-3}$, following a linear warmup for the first 10,000 steps, after which it decays according to a cosine schedule. Training is performed using \texttt{bfloat16} mixed precision for computational efficiency. We use a global batch size of 256 and a context length of 4096. All pre-training is conducted on a cluster of NVIDIA A100 (80GB) GPUs.

\subsubsection{Noise Robustness Analysis}
\label{app:noise_robustness_settings}

To assess the stability of \modelname{} under corrupted inputs, we evaluate noise robustness on the GIFT-Eval benchmark by injecting two types of synthetic perturbations into the input history window. For all experiments, the context length is fixed at 2048 and predictions are generated autoregressively. We fix the random seed at 42 during noise injection and data batching to ensure reproducible comparisons across different noise intensities.

\paragraph{Gaussian Noise.} We perturb the input signal $x$ with zero-mean Gaussian noise scaled by the local standard deviation of each series. The noisy input is defined as $x_{noisy} = x + \epsilon$, where $\epsilon \sim \mathcal{N}(0, (\sigma \cdot \sigma_x)^2)$ and $\sigma_x$ is the standard deviation calculated over the valid segment of the history window. We evaluate five intensity levels where $\sigma \in \{0.0, 0.2, 0.4, 0.6, 0.8\}$. After noise injection, the sequence is re-normalized using its noisy statistics before being fed into the model.

\paragraph{Impulse Noise.} We simulate sudden sensor errors by injecting random spikes into the input sequence. An impulse occurs at any given time step with a probability $p \in \{0.0, 0.05, 0.1, 0.15, 0.2\}$. The magnitude of each impulse is set to $8.0$ times the local standard deviation $\sigma_x$ with the sign of the spike chosen randomly. This setup disrupts the normalization balance and tests model stability against surface-level irregularities.

\paragraph{Evaluation Protocol.} Performance is measured across all 97 configurations of GIFT-Eval. Results are reported as Relative CRPS, calculated by dividing the CRPS of the noisy input by the performance on the clean baseline.

\subsubsection{Representational Analysis and Intervention}
\label{app:representation_settings}

The internal organization and functional controllability of \modelname{} are evaluated through layer-wise probing and latent space steering. These experiments use the small model variant and synthetic signals to isolate fundamental temporal properties.

\paragraph{Layer-wise Feature Probing.}
The linear separability of trend and periodicity features is measured across all Transformer layers and the initial SiGLU embedding layer. Trend probing uses 1,000 pairs of upward and downward linear slopes with random coefficients between 0.5 and 2.0. Periodicity probing uses 1,000 pairs of sine waves with frequencies between 1 and 5 versus white noise. Each sequence has a length of 512 and includes Gaussian noise with $\sigma=0.1$. Hidden states are extracted from the last time step of each layer. Linear separability is quantified using the Linear Discriminant Ratio:
\begin{equation}
    \text{LDR} = \frac{||\mu_1 - \mu_0||^2}{\sigma_1^2 + \sigma_0^2 + \epsilon}
\end{equation}
In this equation $\mu_i$ is the mean vector and $\sigma_i^2$ is the sum of feature-wise variances for class $i$. The term $\epsilon=1e-6$ ensures numerical stability.

\paragraph{Latent Steering and Editing.}
Semantic directions are extracted at the final Transformer layer to intervene in the forecasting process. A concept direction is defined as the difference between class medians in the hidden states of the target layer. Trend steering uses zero-slopes versus linear ramps while periodicity steering uses white noise versus sine waves. The resulting difference vector is normalized to a unit vector $\vec{v}_{unit}$. An energy-proportional injection mechanism is used to ensure fair comparison across different models. Let $E_{model}$ be the average $L_2$-norm of the hidden states in the target layer. The modified hidden state $h'$ is calculated as:
\begin{equation}
    h' = h + (\alpha \cdot E_{model}) \cdot \vec{v}_{unit}
\end{equation}
The injection ratio $\alpha$ is set to 0.2 or 0.5 for synthetic experiments. The intervention is applied via a forward hook during the first step of generation followed by an autoregressive forecast. Inputs are Z-score normalized before steering and denormalized for visualization. This method is validated on real-world time series using the Jena Weather dataset for trend intervention and the Solar dataset for periodicity intervention.

\subsubsection{Scaling Experiments}
\label{app:scaling_settings}

Scaling properties are evaluated by varying the depth and width of the Transformer backbone. Three variants named Small, Base, and Large are implemented. Specific architectural configurations including hidden dimension, intermediate dimension, and the number of layers are scaled as shown in Table~\ref{tab:scaling_params}. 

\begin{table}[ht]
\centering
\caption{Architectural parameters for \modelname{} scaling variants.}
\label{tab:scaling_params}
\begin{tabular}{lccccc}
\toprule
\textbf{Variant} & \textbf{Hidden Size} & \textbf{Intermediate Size} & \textbf{Layers} & \textbf{Heads (Q/KV)} & \textbf{Parameters} \\
\midrule
\modelname{} Small  & 384 & 1024 & 6  & 12 / 12 & 12.7M \\
\modelname{} Base   & 384 & 1024 & 12 & 12 / 12 & 23.3M \\
\modelname{} Large  & 768 & 2048 & 12 & 12 / 12 & 91.7M \\
\bottomrule
\end{tabular}
\end{table}

The number of attention heads and key value heads is fixed at 12 across all scales to ensure a consistent head dimension. Each model is trained with a global batch size of 256 and a context length of 4096.

\subsection{Pre-Training Data Corpus}
\label{pretraining_data_corpus}

The pre-training corpus comprises a balanced mixture of real-world and synthetic time series, sampled at a 1:1 ratio to ensure diversity.

\paragraph{Real-World Data.}
We aggregate time series from two primary repositories: the Chronos training corpus\footnote{\url{https://HuggingFace.co/datasets/autogluon/chronos_datasets}} \citep{ansari2024chronos} and the GiftEval pre-training repository\footnote{\url{https://HuggingFace.co/datasets/Salesforce/GiftEvalPretrain}} \citep{aksu2024gift}. To prevent data leakage, we strictly filter these datasets to ensure zero overlap with the downstream evaluation benchmarks, excluding datasets such as Favorita Sales and KDD Cup 2022. The specific datasets, along with their domains and frequencies, are detailed in Table~\ref{tab:chronos_data} and Table~\ref{tab:gifteval_data}.

To enhance data diversity, we apply the TSMixup augmentation strategy \citep{ansari2024chronos} on the real-world data. TSMixup generates new training samples via a convex combination of $k$ randomly sampled time series segments. Formally, a mixed sample $\mathbf{x}_{\text{mix}} \in \mathbb{R}^L$ is generated as:
\begin{equation}
    \mathbf{x}_{\text{mix}} = \sum_{i=1}^{k} \lambda_i \tilde{\mathbf{x}}_i, \quad \boldsymbol{\lambda} \sim \text{Dir}(\alpha),
\end{equation}
where $\tilde{\mathbf{x}}_i$ represents a z-score normalized segment from a source time series, and the mixing weights $\boldsymbol{\lambda}$ are drawn from a symmetric Dirichlet distribution. We sample $k$ uniformly from $\{1, 2, 3\}$ with a concentration parameter $\alpha=1.5$. Unlike Chronos \citep{ansari2024chronos}, which samples the segment length uniformly from $[128, 2048]$, we fix the training sequence length at $L=4096$. This configuration exposes the model to longer temporal dependencies during pre-training.

\begin{table}[h!]
    \centering
    \caption{Datasets sourced from the Chronos training corpus. }
    \label{tab:chronos_data}
    \begin{tabular}{llrrr}
    \toprule
    \textbf{Dataset Name} & \textbf{Domain} & \textbf{Frequency} & \textbf{\# Series} & \textbf{Avg. Length} \\
    \midrule
    Brazilian Cities Temperature & Nature & Monthly & 12 & 757 \\
    London Smart Meters & Energy & 30 min & 5,560 & 29,951 \\
    Mexico City Bikes & Transport & Hourly & 494 & 78,313 \\
    Pedestrian Counts & Transport & Hourly & 66 & 47,459 \\
    Rideshare & Transport & Hourly & 2,340 & 541 \\
    Solar (5 Min.) & Energy & 5 min & 5,166 & 105,120 \\
    Spanish Energy and Weather & Energy & Hourly & 66 & 35,064 \\
    Taxi (30 Min.) & Transport & 30 min & 2,428 & 1,478 \\
    Taxi (Hourly) & Transport & Hourly & 2,428 & 739 \\
    Uber TLC (Daily) & Transport & Daily & 262 & 181 \\
    Uber TLC (Hourly) & Transport & Hourly & 262 & 4,344 \\
    USHCN & Nature & Daily & 6,090 & 38,653 \\
    Weatherbench (Daily) & Nature & Daily & 225,280 & 14,609 \\
    Weatherbench (Hourly) & Nature & Hourly & 225,280 & 350,639 \\
    Weatherbench (Weekly) & Nature & Weekly & 225,280 & 2,087 \\
    Wiki Daily (100k) & Web & Daily & 100,000 & 2,741 \\
    Wind Farms (Daily) & Energy & Daily & 337 & 354 \\
    Wind Farms (Hourly) & Energy & Hourly & 337 & 8,514 \\
    \bottomrule
    \end{tabular}%
\end{table}

\begin{table}[h!]
    \centering
    \caption{Datasets sourced from the GiftEval pre-training corpus.}
    \label{tab:gifteval_data}
    \begin{tabular}{llrrr}
    \toprule
    \textbf{Dataset Name} & \textbf{Domain} & \textbf{Frequency} & \textbf{\# Series} & \textbf{Avg. Length} \\
    \midrule
    Alibaba Cluster Trace 2018 & Cloud Ops & 5 min & 58,409 & 1,630 \\
    Azure VM Traces 2017 & Cloud Ops & 5 min & 159,472 & 5,553 \\
    BDG-2 Bear & Energy & Hourly & 91 & 16,289 \\
    BDG-2 Fox & Energy & Hourly & 135 & 17,219 \\
    BDG-2 Panther & Energy & Hourly & 105 & 8,760 \\
    BDG-2 Rat & Energy & Hourly & 280 & 16,887 \\
    Beijing Air Quality & Nature & Hourly & 12 & 35,064 \\
    Beijing Subway & Transport & 30 min & 276 & 1,572 \\
    Borealis & Energy & 5 min & 15 & 5,551 \\
    Borg Cluster Data 2011 & Cloud Ops & 5 min & 143,386 & 3,749 \\
    Buildings900K & Energy & Hourly & 1,795,256 & 8,761 \\
    CDC Fluview ILINet & Healthcare & Weekly & 75 & 852 \\
    CDC Fluview WHO NREVSS & Healthcare & Weekly & 74 & 564 \\
    China Air Quality & Nature & Hourly & 437 & 13,133 \\
    GoDaddy & Econ/Fin & Monthly & 3,135 & 41 \\
    HZMetro & Transport & 15 min & 80 & 2,377 \\
    Ideal & Energy & Hourly & 217 & 5,785 \\
    Kaggle Web Traffic Weekly & Web & Weekly & 145,063 & 114 \\
    LargeST 2017 & Transport & 5 min & 8,196 & 105,120 \\
    LargeST 2018 & Transport & 5 min & 8,428 & 105,120 \\
    LargeST 2019 & Transport & 5 min & 8,600 & 105,120 \\
    LargeST 2020 & Transport & 5 min & 8,561 & 105,408 \\
    LargeST 2021 & Transport & 5 min & 8,548 & 105,120 \\
    Los Loop & Transport & 5 min & 207 & 34,272 \\
    Low Carbon London & Energy & Hourly & 713 & 13,385 \\
    PEMS03 & Transport & 5 min & 358 & 26,208 \\
    PEMS04 & Transport & 5 min & 307 & 16,992 \\
    PEMS07 & Transport & 5 min & 883 & 28,224 \\
    PEMS08 & Transport & 5 min & 170 & 17,856 \\
    PEMS Bay & Transport & 5 min & 325 & 52,128 \\
    Q-Traffic & Transport & 15 min & 45,148 & 5,856 \\
    Residential Load Power & Energy & Minutely & 271 & 538,725 \\
    Residential PV Power & Energy & Minutely & 233 & 537,935 \\
    Sceaux & Energy & Hourly & 1 & 34,223 \\
    SHMetro & Transport & 15 min & 288 & 8,809 \\
    SMART & Energy & Hourly & 5 & 19,142 \\
    Solar Power & Energy & 4 sec & 1 & 7,397,222 \\
    Subseasonal & Climate & Daily & 862 & 16,470 \\
    Subseasonal Precipitation & Climate & Daily & 862 & 11,323 \\
    Wind Power & Energy & 4 sec & 1 & 7,397,147 \\
    \bottomrule
    \end{tabular}%
\end{table}

\paragraph{Synthetic Data.}
To scale the training data and introduce complex temporal dynamics, we utilize the CauKer generator \citep{xie2025cauker}. CauKer constructs multivariate time series via a Structural Causal Model where dependencies are encoded in a random Directed Acyclic Graph. The generative process consists of sampling root nodes from Gaussian Processes using composite kernels and stochastic mean functions, followed by generating child nodes through edge-specific non-linear activation functions applied to their parents and subsequent random linear aggregation. This mechanism allows the synthesis of diverse structural patterns such as seasonality, trend shifts, and non-Gaussian dependencies without requiring real-world semantic labels. We generated a total of 15.7 million univariate time series, each with a fixed sequence length of $L=4096$, using the default hyperparameter configuration described in \citet{moroshan2025tempopfn}.

\subsection{Benchmarks and Metrics}
\label{app:benchmarks_and_metrics}

We evaluate the zero-shot forecasting performance of \modelname{} on two standardized benchmarks to ensure a rigorous assessment across diverse data distributions and forecasting horizons. 

\subsubsection{Evaluation Benchmarks}

\textbf{GIFT-Eval Benchmark} \citep{aksu2024gift}. 
GIFT-Eval serves as our primary benchmark for evaluating general-purpose forecasting performance. It consists of 23 datasets spanning seven application domains, including energy, transportation, and finance, and covers ten temporal frequencies ranging from secondly to yearly data. By combining datasets, frequencies, and forecast lengths, GIFT-Eval defines 97 distinct evaluation settings that systematically assess short-, medium-, and long-term forecasting horizons. All evaluations are conducted in a strict zero-shot setting, where the target time series in the test split are excluded from the pre-training corpus.

\textbf{fev-bench} \citep{shchur2025fev}. 
fev-bench is a realistic forecasting benchmark designed to reflect practical scenarios involving multivariate targets and exogenous covariates. It contains 100 forecasting tasks across seven domains, with 46 tasks explicitly incorporating dynamic or static covariates. Evaluation employs a rolling-origin protocol, averaging performance over multiple temporal windows to enhance statistical reliability.

\subsubsection{Evaluation Metrics}
\textbf{Mean Absolute Scaled Error (MASE).}
MASE evaluates point forecasting accuracy by scaling the mean absolute error relative to the in-sample mean absolute error of a naive seasonal baseline. For a forecast horizon $l$, MASE is defined as:
\begin{equation}
\mathrm{MASE}
=
\frac{
\frac{1}{l}\sum_{t=1}^{l} |y_t - \hat{y}_t|
}{
\frac{1}{T-m}\sum_{t=m+1}^{T} |y_t - y_{t-m}|
},
\end{equation}
where $y_t$ is the ground truth, $\hat{y}_t$ is the prediction, $m$ denotes the seasonal period and $T$ is the length of the training series. Lower values indicate better forecasting accuracy.
MASE is used as the primary point forecasting metric in both benchmarks.

\textbf{Continuous Ranked Probability Score (CRPS).}
CRPS measures the quality of probabilistic forecasts by comparing the predicted cumulative distribution function $F$ with the observed outcome $y$:
\begin{equation}
\mathrm{CRPS}(F, y)
=
\int_{-\infty}^{\infty}
\big(F(z) - \mathbb{I}(z \ge y)\big)^2 \, dz,
\end{equation}
where $\mathbb{I}(\cdot)$ is the indicator function. In practice, CRPS is approximated by averaging the quantile loss over a set of equidistant quantile levels. It serves as the primary probabilistic metric for GIFT-Eval.

\textbf{Weighted Quantile Loss (WQL).}
WQL evaluates probabilistic forecasts based on a weighted aggregation of quantile losses across a set of quantile levels $Q$. The metric is normalized by the total absolute value of the target series to ensure scale independence:
\begin{equation}
\mathrm{WQL} = \frac{\sum_{t=1}^{l} \sum_{q \in Q} 2 \cdot \max \left( q(y_t - \hat{y}_t^{(q)}), \, (1-q)(\hat{y}_t^{(q)} - y_t) \right)}{\sum_{t=1}^{l} |y_t|},
\end{equation}
where $\hat{y}_t^{(q)}$ is the predicted $q$-th quantile at step $t$. WQL penalizes both under- and over-estimation in a distribution-aware manner and constitutes the primary probabilistic evaluation metric for fev-bench.

\subsubsection{Metric Aggregation}
To aggregate performance across diverse datasets with varying scales, we report the geometric mean of relative metrics. For each evaluation setting and metric, we first normalize the model's score by the score of a Seasonal Naive baseline. We then compute the geometric mean of these relative scores across all tasks.

\section{Extended Results}
\label{app:extended_results}

\subsection{Benchmark Results}
\subsubsection{GIFT-Eval}

To complement the aggregated zero-shot results reported in Section~\ref{sec:zs-results}, we present detailed performance metrics for the full GIFT-Eval benchmark. Table~\ref{tab:full_gift_mase} reports MASE scores across all 97 evaluation configurations, while Table~\ref{tab:full_gift_crps} provides the corresponding CRPS results.

\begin{scriptsize}
\renewcommand{\arraystretch}{1.1}
\begin{longtable}{lccccc}
  \caption{Detailed MASE scores of different zero-shot models on the GIFT-Eval benchmark.} \\
    \hline \textbf{Dataset} & \multicolumn{1}{l}{\textbf{\modelname{}}} & \multicolumn{1}{l}{\textbf{Chronos-2}} & \multicolumn{1}{l}{\textbf{Moirai2}} & \multicolumn{1}{l}{\textbf{Sundial base}} & \multicolumn{1}{l}{\textbf{Yinglong 300M}} \\  \hline 
\endfirsthead
\multicolumn{6}{c}%
{{\bfseries \tablename\ \thetable{} -- continued from previous page}} \\
\hline \textbf{Dataset} & \multicolumn{1}{l}{\textbf{\modelname{}}} & \multicolumn{1}{l}{\textbf{Chronos-2}} & \multicolumn{1}{l}{\textbf{Moirai2}} & \multicolumn{1}{l}{\textbf{Sundial base}} & \multicolumn{1}{l}{\textbf{Yinglong 300M}} \\  \hline  
\endhead

\hline 
\endfoot

\endlastfoot
    bitbrains\_fast\_storage/5T/long & 0.9896  & 0.8712  & 0.9175  & 1.0111  & 1.0096  \\
    bitbrains\_fast\_storage/5T/medium & 1.0385  & 0.9588  & 0.9918  & 1.1079  & 1.0718  \\
    bitbrains\_fast\_storage/5T/short & 0.7565  & 0.6564  & 0.6865  & 0.7401  & 0.8032  \\
    bitbrains\_fast\_storage/H/short & 1.1247  & 0.9310  & 1.0675  & 1.1499  & 1.1160  \\
    bitbrains\_rnd/5T/long & 3.4559  & 3.2754  & 3.3334  & 3.5225  & 3.4700  \\
    bitbrains\_rnd/5T/medium & 4.4881  & 4.3592  & 4.3997  & 4.5622  & 4.4983  \\
    bitbrains\_rnd/5T/short & 1.7262  & 1.6181  & 1.6521  & 1.7147  & 1.7857  \\
    bitbrains\_rnd/H/short & 5.9095  & 5.8047  & 5.8093  & 5.9797  & 5.8916  \\
    bizitobs\_application/10S/long & 3.4265  & 2.9497  & 3.4279  & 3.7052  & 4.5998  \\
    bizitobs\_application/10S/medium & 2.9177  & 1.8026  & 2.4790  & 2.8565  & 3.8681  \\
    bizitobs\_application/10S/short & 1.5486  & 1.0039  & 1.4386  & 1.4287  & 1.8178  \\
    bizitobs\_l2c/5T/long & 1.0592  & 0.6472  & 0.5799  & 0.6350  & 1.2143  \\
    bizitobs\_l2c/5T/medium & 0.7909  & 0.5826  & 0.5360  & 0.5302  & 0.8765  \\
    bizitobs\_l2c/5T/short & 0.2922  & 0.2665  & 0.2906  & 0.2481  & 0.2864  \\
    bizitobs\_l2c/H/long & 0.5562  & 0.5713  & 0.6108  & 0.6650  & 0.8683  \\
    bizitobs\_l2c/H/medium & 0.4752  & 0.4980  & 0.5086  & 0.5502  & 0.7066  \\
    bizitobs\_l2c/H/short & 0.4661  & 0.4126  & 0.5027  & 0.4764  & 0.5545  \\
    bizitobs\_service/10S/long & 1.5718  & 1.3256  & 1.3712  & 1.4573  & 2.2091  \\
    bizitobs\_service/10S/medium & 1.3618  & 0.9790  & 1.1320  & 1.2721  & 2.0245  \\
    bizitobs\_service/10S/short & 0.9283  & 0.7169  & 0.8739  & 0.8388  & 1.1376  \\
    car\_parts/M/short & 0.8479  & 0.8360  & 0.8266  & 0.9568  & 1.0652  \\
    covid\_deaths/D/short & 47.5906  & 32.5392  & 36.9581  & 60.3751  & 45.4044  \\
    electricity/15T/long & 0.9178  & 0.8554  & 0.9071  & 0.9062  & 0.9184  \\
    electricity/15T/medium & 0.8724  & 0.8041  & 0.8413  & 0.8541  & 0.8830  \\
    electricity/15T/short & 1.1189  & 0.9223  & 0.8698  & 0.8950  & 1.0721  \\
    electricity/D/short & 1.4696  & 1.4149  & 1.4216  & 1.4559  & 1.3961  \\
    electricity/H/long & 1.2876  & 1.1925  & 1.2622  & 1.0754  & 1.2904  \\
    electricity/H/medium & 1.1481  & 1.0692  & 1.0840  & 0.9935  & 1.1394  \\
    electricity/H/short & 0.9498  & 0.9779  & 0.8868  & 0.9319  & 1.0921  \\
    electricity/W/short & 1.7262  & 1.3851  & 1.6595  & 1.6145  & 1.5986  \\
    ett1/15T/long & 1.0855  & 1.0134  & 1.0669  & 1.0883  & 1.0331  \\
    ett1/15T/medium & 1.0516  & 0.9860  & 1.0390  & 1.0667  & 1.0285  \\
    ett1/15T/short & 0.6873  & 0.6869  & 0.6870  & 0.7097  & 0.7168  \\
    ett1/D/short & 1.8616  & 1.6403  & 1.7477  & 1.9034  & 1.7279  \\
    ett1/H/long & 1.4208  & 1.3805  & 1.4650  & 1.4054  & 1.3701  \\
    ett1/H/medium & 1.2494  & 1.2738  & 1.3043  & 1.2876  & 1.2603  \\
    ett1/H/short & 0.8020  & 0.7868  & 0.8362  & 0.8295  & 0.8324  \\
    ett1/W/short & 1.5872  & 1.5901  & 1.4887  & 1.8430  & 1.5891  \\
    ett2/15T/long & 0.9200  & 0.8903  & 0.9591  & 0.9212  & 0.9057  \\
    ett2/15T/medium & 0.9101  & 0.8487  & 0.9196  & 0.9072  & 0.8876  \\
    ett2/15T/short & 0.7802  & 0.7262  & 0.7664  & 0.7473  & 0.7527  \\
    ett2/D/short & 1.3482  & 1.2908  & 1.3029  & 1.5071  & 1.2995  \\
    ett2/H/long & 1.1140  & 1.0480  & 1.0515  & 1.1394  & 1.0566  \\
    ett2/H/medium & 0.9994  & 1.0902  & 1.0853  & 1.1150  & 1.0184  \\
    ett2/H/short & 0.7242  & 0.7363  & 0.7354  & 0.7705  & 0.7407  \\
    ett2/W/short & 0.8048  & 0.7716  & 0.8076  & 0.9365  & 0.9203  \\
    hierarchical\_sales/D/short & 0.7488  & 0.7440  & 0.7480  & 0.7895  & 0.7628  \\
    hierarchical\_sales/W/short & 0.7273  & 0.7093  & 0.7381  & 0.7510  & 0.7471  \\
    hospital/M/short & 0.8353  & 0.7399  & 0.7650  & 0.8374  & 0.7926  \\
    jena\_weather/10T/long & 0.6793  & 0.6919  & 0.7331  & 0.6790  & 0.6389  \\
    jena\_weather/10T/medium & 0.6394  & 0.6057  & 0.6510  & 0.6386  & 0.6169  \\
    jena\_weather/10T/short & 0.2945  & 0.2650  & 0.3154  & 0.2972  & 0.3190  \\
    jena\_weather/D/short & 0.9983  & 1.0823  & 1.0152  & 0.9311  & 1.1224  \\
    jena\_weather/H/long & 1.0524  & 1.0084  & 1.0389  & 1.0882  & 1.0298  \\
    jena\_weather/H/medium & 0.9410  & 0.7857  & 0.8173  & 0.8685  & 0.8865  \\
    jena\_weather/H/short & 0.5324  & 0.5226  & 0.5365  & 0.5388  & 0.5433  \\
    kdd\_cup\_2018/D/short & 1.1772  & 1.1978  & 1.2415  & 1.1736  & 1.1829  \\
    kdd\_cup\_2018/H/long & 1.0783  & 1.0469  & 1.0698  & 0.7747  & 1.0039  \\
    kdd\_cup\_2018/H/medium & 1.1004  & 1.0401  & 1.1240  & 0.8408  & 1.0341  \\
    kdd\_cup\_2018/H/short & 0.9285  & 0.9467  & 1.0065  & 0.8006  & 0.9270  \\
    loop\_seattle/5T/long & 0.9265  & 0.9113  & 0.9282  & 0.8928  & 1.0697  \\
    loop\_seattle/5T/medium & 0.8437  & 0.8455  & 0.8646  & 0.8196  & 1.0226  \\
    loop\_seattle/5T/short & 0.5444  & 0.5407  & 0.5418  & 0.5419  & 0.6071  \\
    loop\_seattle/D/short & 0.8916  & 0.8878  & 0.8936  & 0.8996  & 0.9073  \\
    loop\_seattle/H/long & 0.9127  & 0.8735  & 0.9038  & 0.9871  & 0.9806  \\
    loop\_seattle/H/medium & 0.9260  & 0.9076  & 0.9398  & 1.0140  & 0.9658  \\
    loop\_seattle/H/short & 0.8701  & 0.8250  & 0.8503  & 0.8804  & 0.8947  \\
    m\_dense/D/short & 0.7159  & 0.6151  & 0.6916  & 0.6813  & 0.7452  \\
    m\_dense/H/long & 0.7500  & 0.6824  & 0.7182  & 0.7709  & 0.8426  \\
    m\_dense/H/medium & 0.7266  & 0.6971  & 0.7039  & 0.7587  & 0.7876  \\
    m\_dense/H/short & 0.8036  & 0.7817  & 0.7930  & 0.7912  & 0.9291  \\
    m4\_daily/D/short & 3.6707  & 3.4193  & 3.0710  & 3.7149  & 3.5132  \\
    m4\_hourly/H/short & 0.7099  & 0.7884  & 0.8085  & 0.8693  & 0.9250  \\
    m4\_monthly/M/short & 1.0844  & 0.9032  & 0.9396  & 1.0992  & 1.0482  \\
    m4\_quarterly/Q/short & 1.4266  & 1.1543  & 1.1834  & 1.4574  & 1.3902  \\
    m4\_weekly/W/short & 2.1198  & 2.0263  & 2.1235  & 2.3961  & 2.2479  \\
    m4\_yearly/A/short & 4.6259  & 3.2362  & 3.3204  & 4.3299  & 4.3125  \\
    restaurant/D/short & 0.7028  & 0.6777  & 0.6960  & 0.7043  & 0.7029  \\
    saugeen/D/short & 2.9006  & 2.9834  & 2.7014  & 2.7832  & 3.1355  \\
    saugeen/M/short & 0.7349  & 0.7235  & 0.7310  & 0.7528  & 0.7851  \\
    saugeen/W/short & 1.2244  & 1.1876  & 1.4042  & 1.1983  & 1.1860  \\
    solar/10T/long & 1.0106  & 0.7941  & 1.0108  & 0.9494  & 0.8863  \\
    solar/10T/medium & 0.9337  & 0.7900  & 0.9995  & 0.9408  & 0.8896  \\
    solar/10T/short & 0.9534  & 0.7784  & 0.6575  & 0.8373  & 1.1073  \\
    solar/D/short & 0.9700  & 0.9610  & 1.0639  & 1.0813  & 0.9718  \\
    solar/H/long & 0.9104  & 0.9951  & 0.8834  & 0.7493  & 0.9605  \\
    solar/H/medium & 0.8704  & 1.0694  & 0.8482  & 0.7667  & 0.9668  \\
    solar/H/short & 0.8441  & 0.9845  & 0.8794  & 0.7869  & 0.9114  \\
    solar/W/short & 1.0621  & 1.0123  & 1.1967  & 0.9809  & 1.7994  \\
    sz\_taxi/15T/long & 0.5179  & 0.5097  & 0.5303  & 0.5372  & 0.5113  \\
    sz\_taxi/15T/medium & 0.5405  & 0.5346  & 0.5520  & 0.5632  & 0.5409  \\
    sz\_taxi/15T/short & 0.5413  & 0.5429  & 0.5464  & 0.5538  & 0.5506  \\
    sz\_taxi/H/short & 0.5848  & 0.5603  & 0.5633  & 0.5814  & 0.5678  \\
    temperature\_rain/D/short & 1.3653  & 1.3102  & 1.3483  & 1.4299  & 1.4169  \\
    us\_births/D/short & 0.3712  & 0.3266  & 0.3733  & 0.3887  & 0.5056  \\
    us\_births/M/short & 0.9707  & 0.6328  & 0.8113  & 1.1581  & 0.7297  \\
    us\_births/W/short & 1.1293  & 0.9790  & 0.8714  & 1.3624  & 1.2366  \\
    \bottomrule
  \label{tab:full_gift_mase}%
\end{longtable}%
\end{scriptsize}

\begin{scriptsize}
\renewcommand{\arraystretch}{1.1}
\begin{longtable}{lccccc}
  \caption{Detailed CRPS scores of different zero-shot models on the GIFT-Eval benchmark.} \\
    \hline \textbf{Dataset} & \multicolumn{1}{l}{\textbf{\modelname{}}} & \multicolumn{1}{l}{\textbf{Chronos-2}} & \multicolumn{1}{l}{\textbf{Moirai2}} & \multicolumn{1}{l}{\textbf{Sundial base}} & \multicolumn{1}{l}{\textbf{Yinglong 300M}} \\  \hline 
\endfirsthead
\multicolumn{6}{c}%
{{\bfseries \tablename\ \thetable{} -- continued from previous page}} \\
\hline \textbf{Dataset} & \multicolumn{1}{l}{\textbf{\modelname{}}} & \multicolumn{1}{l}{\textbf{Chronos-2}} & \multicolumn{1}{l}{\textbf{Moirai2}} & \multicolumn{1}{l}{\textbf{Sundial base}} & \multicolumn{1}{l}{\textbf{Yinglong 300M}} \\  \hline  
\endhead

\hline 
\endfoot

\endlastfoot
    bitbrains\_fast\_storage/5T/long & 1.1176  & 0.7026  & 0.8066  & 0.8113  & 0.7091  \\
    bitbrains\_fast\_storage/5T/medium & 0.8021  & 0.6233  & 0.6939  & 0.7276  & 0.6452  \\
    bitbrains\_fast\_storage/5T/short & 0.4149  & 0.3906  & 0.4275  & 0.4617  & 0.4236  \\
    bitbrains\_fast\_storage/H/short & 0.7467  & 0.6635  & 0.6149  & 0.7644  & 0.6315  \\
    bitbrains\_rnd/5T/long & 0.5814  & 0.8730  & 0.5698  & 0.7152  & 0.6887  \\
    bitbrains\_rnd/5T/medium & 0.7270  & 1.0046  & 0.5961  & 0.7303  & 0.6524  \\
    bitbrains\_rnd/5T/short & 0.4162  & 0.4157  & 0.4043  & 0.4334  & 0.4255  \\
    bitbrains\_rnd/H/short & 0.6217  & 0.8035  & 0.6697  & 0.7254  & 0.6726  \\
    bizitobs\_application/10S/long & 0.0574  & 0.0454  & 0.0565  & 0.0614  & 0.0606  \\
    bizitobs\_application/10S/medium & 0.0484  & 0.0257  & 0.0369  & 0.0457  & 0.0483  \\
    bizitobs\_application/10S/short & 0.0164  & 0.0096  & 0.0130  & 0.0163  & 0.0168  \\
    bizitobs\_l2c/5T/long & 0.5973  & 0.2978  & 0.3001  & 0.3101  & 0.5756  \\
    bizitobs\_l2c/5T/medium & 0.3911  & 0.2472  & 0.2606  & 0.2341  & 0.3788  \\
    bizitobs\_l2c/5T/short & 0.0776  & 0.0691  & 0.0844  & 0.0670  & 0.0770  \\
    bizitobs\_l2c/H/long & 0.2686  & 0.2673  & 0.3213  & 0.3248  & 0.4065  \\
    bizitobs\_l2c/H/medium & 0.2420  & 0.2359  & 0.2738  & 0.2763  & 0.3300  \\
    bizitobs\_l2c/H/short & 0.2002  & 0.1765  & 0.2347  & 0.2228  & 0.2294  \\
    bizitobs\_service/10S/long & 0.0583  & 0.0511  & 0.0539  & 0.0568  & 0.0621  \\
    bizitobs\_service/10S/medium & 0.0462  & 0.0221  & 0.0339  & 0.0440  & 0.0455  \\
    bizitobs\_service/10S/short & 0.0184  & 0.0101  & 0.0136  & 0.0163  & 0.0168  \\
    car\_parts/M/short & 1.0003  & 0.9660  & 0.9362  & 1.1888  & 1.1910  \\
    covid\_deaths/D/short & 0.0653  & 0.0348  & 0.0281  & 0.1311  & 0.0783  \\
    electricity/15T/long & 0.0859  & 0.0717  & 0.0832  & 0.0822  & 0.0782  \\
    electricity/15T/medium & 0.0850  & 0.0714  & 0.0800  & 0.0822  & 0.0788  \\
    electricity/15T/short & 0.0936  & 0.0784  & 0.0771  & 0.0836  & 0.0929  \\
    electricity/D/short & 0.0598  & 0.0581  & 0.0541  & 0.0638  & 0.0544  \\
    electricity/H/long & 0.1133  & 0.0878  & 0.0975  & 0.0930  & 0.0974  \\
    electricity/H/medium & 0.0930  & 0.0759  & 0.0804  & 0.0799  & 0.0824  \\
    electricity/H/short & 0.0644  & 0.0679  & 0.0645  & 0.0694  & 0.0783  \\
    electricity/W/short & 0.0671  & 0.0559  & 0.0664  & 0.0720  & 0.0571  \\
    ett1/15T/long & 0.2772  & 0.2411  & 0.2684  & 0.2527  & 0.2344  \\
    ett1/15T/medium & 0.2736  & 0.2330  & 0.2605  & 0.2600  & 0.2433  \\
    ett1/15T/short & 0.1607  & 0.1653  & 0.1601  & 0.1773  & 0.1665  \\
    ett1/D/short & 0.3049  & 0.2742  & 0.2865  & 0.3729  & 0.2842  \\
    ett1/H/long & 0.3031  & 0.2752  & 0.3231  & 0.2827  & 0.2636  \\
    ett1/H/medium & 0.2709  & 0.2606  & 0.2873  & 0.2691  & 0.2519  \\
    ett1/H/short & 0.1734  & 0.1706  & 0.1851  & 0.1902  & 0.1821  \\
    ett1/W/short & 0.2684  & 0.2714  & 0.2488  & 0.4039  & 0.2702  \\
    ett2/15T/long & 0.1055  & 0.0929  & 0.1023  & 0.0977  & 0.0924  \\
    ett2/15T/medium & 0.1031  & 0.0874  & 0.0979  & 0.0958  & 0.0902  \\
    ett2/15T/short & 0.0672  & 0.0622  & 0.0660  & 0.0687  & 0.0656  \\
    ett2/D/short & 0.0950  & 0.0942  & 0.0930  & 0.1026  & 0.0924  \\
    ett2/H/long & 0.1295  & 0.1051  & 0.1086  & 0.1168  & 0.1075  \\
    ett2/H/medium & 0.1071  & 0.1091  & 0.1107  & 0.1136  & 0.1043  \\
    ett2/H/short & 0.0641  & 0.0642  & 0.0642  & 0.0721  & 0.0645  \\
    ett2/W/short & 0.0905  & 0.0896  & 0.0851  & 0.0982  & 0.0913  \\
    hierarchical\_sales/D/short & 0.5816  & 0.5789  & 0.5770  & 0.6492  & 0.5887  \\
    hierarchical\_sales/W/short & 0.3459  & 0.3407  & 0.3521  & 0.3902  & 0.3713  \\
    hospital/M/short & 0.0616  & 0.0510  & 0.0518  & 0.0609  & 0.0572  \\
    jena\_weather/10T/long & 0.0559  & 0.0514  & 0.0596  & 0.0560  & 0.0518  \\
    jena\_weather/10T/medium & 0.0551  & 0.0496  & 0.0587  & 0.0542  & 0.0505  \\
    jena\_weather/10T/short & 0.0293  & 0.0303  & 0.0360  & 0.0308  & 0.0298  \\
    jena\_weather/D/short & 0.0440  & 0.0467  & 0.0434  & 0.0479  & 0.0499  \\
    jena\_weather/H/long & 0.0641  & 0.0586  & 0.0580  & 0.0657  & 0.0602  \\
    jena\_weather/H/medium & 0.0612  & 0.0503  & 0.0550  & 0.0584  & 0.0573  \\
    jena\_weather/H/short & 0.0417  & 0.0419  & 0.0421  & 0.0496  & 0.0452  \\
    kdd\_cup\_2018/D/short & 0.3679  & 0.3666  & 0.3888  & 0.3956  & 0.3738  \\
    kdd\_cup\_2018/H/long & 0.5502  & 0.4453  & 0.5183  & 0.3747  & 0.4394  \\
    kdd\_cup\_2018/H/medium & 0.5113  & 0.4172  & 0.4955  & 0.3771  & 0.4170  \\
    kdd\_cup\_2018/H/short & 0.3755  & 0.3736  & 0.4259  & 0.3515  & 0.3741  \\
    loop\_seattle/5T/long & 0.0902  & 0.0795  & 0.0871  & 0.0845  & 0.0963  \\
    loop\_seattle/5T/medium & 0.0806  & 0.0742  & 0.0803  & 0.0774  & 0.0921  \\
    loop\_seattle/5T/short & 0.0465  & 0.0462  & 0.0463  & 0.0497  & 0.0524  \\
    loop\_seattle/D/short & 0.0431  & 0.0424  & 0.0431  & 0.0471  & 0.0428  \\
    loop\_seattle/H/long & 0.0674  & 0.0593  & 0.0657  & 0.0722  & 0.0680  \\
    loop\_seattle/H/medium & 0.0674  & 0.0628  & 0.0691  & 0.0753  & 0.0668  \\
    loop\_seattle/H/short & 0.0608  & 0.0577  & 0.0618  & 0.0666  & 0.0630  \\
    m\_dense/D/short & 0.0663  & 0.0578  & 0.0701  & 0.0668  & 0.0728  \\
    m\_dense/H/long & 0.1305  & 0.1133  & 0.1196  & 0.1304  & 0.1451  \\
    m\_dense/H/medium & 0.1235  & 0.1172  & 0.1200  & 0.1282  & 0.1340  \\
    m\_dense/H/short & 0.1342  & 0.1283  & 0.1319  & 0.1332  & 0.1565  \\
    m4\_daily/D/short & 0.0237  & 0.0229  & 0.0201  & 0.0266  & 0.0225  \\
    m4\_hourly/H/short & 0.0186  & 0.0232  & 0.0236  & 0.0230  & 0.0248  \\
    m4\_monthly/M/short & 0.1055  & 0.0899  & 0.0947  & 0.1163  & 0.1040  \\
    m4\_quarterly/Q/short & 0.0859  & 0.0732  & 0.0754  & 0.0931  & 0.0858  \\
    m4\_weekly/W/short & 0.0379  & 0.0367  & 0.0400  & 0.0435  & 0.0409  \\
    m4\_yearly/A/short & 0.1590  & 0.1115  & 0.1158  & 0.1600  & 0.1516  \\
    restaurant/D/short & 0.2641  & 0.2543  & 0.2604  & 0.2857  & 0.2661  \\
    saugeen/D/short & 0.3487  & 0.3494  & 0.3284  & 0.3792  & 0.3812  \\
    saugeen/M/short & 0.2892  & 0.2872  & 0.2910  & 0.3320  & 0.3278  \\
    saugeen/W/short & 0.3708  & 0.3469  & 0.4193  & 0.4058  & 0.3604  \\
    solar/10T/long & 0.3953  & 0.2901  & 0.4084  & 0.3648  & 0.3512  \\
    solar/10T/medium & 0.3780  & 0.2902  & 0.4124  & 0.3726  & 0.3482  \\
    solar/10T/short & 0.4855  & 0.3860  & 0.3504  & 0.4444  & 0.5534  \\
    solar/D/short & 0.2789  & 0.2666  & 0.3030  & 0.3237  & 0.2781  \\
    solar/H/long & 0.3640  & 0.3397  & 0.3319  & 0.2930  & 0.3519  \\
    solar/H/medium & 0.3547  & 0.3630  & 0.3329  & 0.3092  & 0.3742  \\
    solar/H/short & 0.3352  & 0.3462  & 0.3424  & 0.3286  & 0.3554  \\
    solar/W/short & 0.1434  & 0.1387  & 0.1627  & 0.1478  & 0.2550  \\
    sz\_taxi/15T/long & 0.2266  & 0.2011  & 0.2167  & 0.2211  & 0.1984  \\
    sz\_taxi/15T/medium & 0.2245  & 0.2020  & 0.2136  & 0.2279  & 0.2034  \\
    sz\_taxi/15T/short & 0.1996  & 0.2001  & 0.2014  & 0.2233  & 0.2030  \\
    sz\_taxi/H/short & 0.1415  & 0.1353  & 0.1360  & 0.1540  & 0.1367  \\
    temperature\_rain/D/short & 0.5531  & 0.5388  & 0.5608  & 0.6200  & 0.5710  \\
    us\_births/D/short & 0.0190  & 0.0165  & 0.0196  & 0.0216  & 0.0262  \\
    us\_births/M/short & 0.0206  & 0.0125  & 0.0169  & 0.0283  & 0.0148  \\
    us\_births/W/short & 0.0132  & 0.0114  & 0.0108  & 0.0173  & 0.0147  \\
    \bottomrule
  \label{tab:full_gift_crps}%
\end{longtable}%
\end{scriptsize}

\subsubsection{Fev-bench}
\label{app:fev-bench}

\begin{figure*}[htbp]
    \centering
    \includegraphics[width=\textwidth]{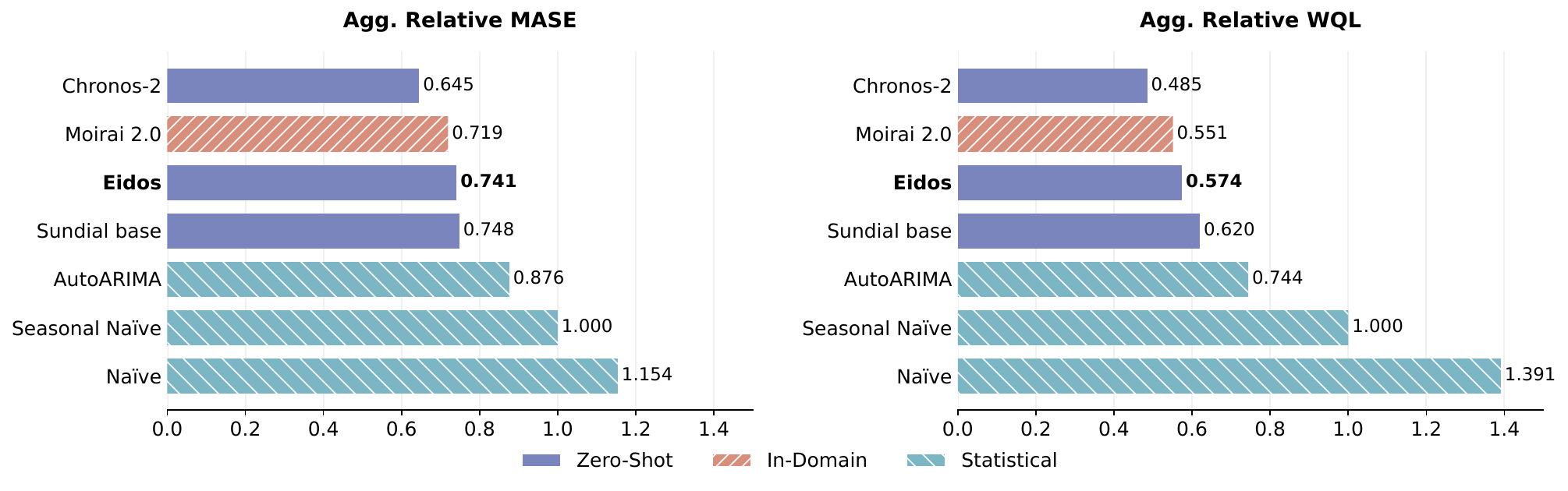}
    \caption{Performance on the fev-bench across 100 evaluation tasks. Results are presented as Aggregated Relative MASE and WQL where lower values indicate better accuracy.}
    \label{fig:fev_bench}
\end{figure*}

We further extend our evaluation to fev-bench, a rigorous benchmark comprising 100 diverse forecasting tasks. As illustrated in Figure~\ref{fig:fev_bench}, \modelname{} demonstrates strong zero-shot performance, achieving a Normalized MASE of 0.741 and Normalized WQL of 0.574. In this setting, our model consistently outperforms Sundial and traditional statistical methods, including AutoARIMA and Seasonal Na\"ive, supporting its suitability as a general-purpose forecaster.

While \modelname{} remains highly competitive, the results also reflect specific characteristics of the fev-bench protocol that differ from GIFT-Eval. A significant proportion of tasks in fev-bench involve dynamic covariates and multivariate targets, which are better aligned with models such as Chronos-2 that explicitly support multivariate attention and covariate integration. In contrast, our current evaluation focuses strictly on univariate zero-shot forecasting without auxiliary features. Furthermore, the evaluation tasks in fev-bench are predominantly skewed toward short- and medium-term horizons. As discussed in Section~\ref{sec:ablation-results}, the design emphasis of \modelname{}'s latent-space predictive learning is on capturing coherent long-term dynamics and structural stability. In short-horizon regimes, observation-space models often perform well by closely matching recent surface-level patterns, whereas the benefits of a structured latent manifold become more apparent as the prediction window extends. Despite these characteristics of the benchmark, \modelname{} maintains competitive performance, highlighting the robustness of the learned representations across evaluation protocols.

\subsection{Model Efficiency}
\label{app:model_efficiency}

\begin{table}[htbp]
  \centering
  \caption{Speed–parameter count comparison.}
    \begin{tabular}{lcc}
    \toprule
       & Model size (M) & Inference time (ms) \\
    \midrule
    \modelname{} & 12.65 & 6.71 \\
    Chronos-2 & 119.48 & 26.38 \\
    Moirai2 & 11.39 & 28.8 \\
    Sundial base & 128.33 & 80.22 \\
    \bottomrule
    \end{tabular}%
  \label{tab:model_efficiency}%
\end{table}%

Computational efficiency is evaluated by comparing model parameter counts and inference latency. Table~\ref{tab:model_efficiency} summarizes metrics measured on an NVIDIA A100 GPU using a batch size of 1 with a context length of 512 and a prediction horizon of 96. \modelname{} achieves a latency of 6.71 ms. This performance is substantially faster than larger models like Chronos-2 and Sundial base. Although Moirai2 has a parameter count comparable to \modelname{}, its inference latency is more than four times higher. These results indicate that the compact architecture of \modelname{} provides a superior balance between computational speed and forecasting accuracy. This efficiency supports the zero-shot performance discussed in Section~\ref{sec:zs-results} and makes the model suitable for time sensitive deployment.

\subsection{Noise Robustness Analysis}

\begin{table}[htbp]
  \centering
  \caption{Normalized CRPS on GIFT-Eval under increasing levels of additive Gaussian noise.}
    \begin{tabular}{rccccc}
    \toprule
    Noise Level & 0  & 0.2 & 0.4 & 0.6 & 0.8 \\
    \midrule
    \modelname{} & 0.547 & 0.5813 & 0.6205 & 0.6676 & 0.7199 \\
    Chronos-2 & 0.4807 & 0.5226 & 0.5834 & 0.6525 & 0.7248 \\
    Moirai2 & 0.5153 & 0.5812 & 0.6572 & 0.7236 & 0.7967 \\
    \bottomrule
    \end{tabular}%
  \label{tab:robust_gausian}%
\end{table}%

\begin{table}[htbp]
  \centering
  \caption{Normalized CRPS on GIFT-Eval under varying probabilities of impulse noise.}
    \begin{tabular}{rccccc}
    \toprule
    Noise Level & 0  & 0.05 & 0.1 & 0.15 & 0.2 \\
    \midrule
    \modelname{} & 0.547 & 0.6031 & 0.6794 & 0.7818 & 0.9253 \\
    Chronos-2 & 0.4807 & 0.538 & 0.6346 & 0.7822 & 0.986 \\
    Moirai2 & 0.5153 & 0.6356 & 0.7551 & 0.9099 & 1.0742 \\
    \bottomrule
    \end{tabular}%
  \label{tab:robust_impulse}%
\end{table}%

This section reports the detailed results of the noise robustness experiments described in Section~\ref{sec:noise_robustness_analysis}. We evaluate model robustness on the GIFT-Eval benchmark by injecting controlled perturbations into the input history, including additive Gaussian noise and impulse noise.

Table~\ref{tab:robust_gausian} presents the Normalized CRPS results under increasing levels of additive Gaussian noise, scaled by the standard deviation of each time series. While Chronos-2 achieves lower error on clean inputs, its performance degrades more rapidly as noise intensity increases. At the highest noise level ($0.8$), \modelname{} attains lower CRPS than Chronos-2, indicating stronger robustness under severe stochastic perturbations.

Table~\ref{tab:robust_impulse} reports the results under impulse noise, implemented as random spike bursts with varying probabilities. A similar pattern is observed across models. In particular, at higher noise probabilities ($0.15$ and $0.2$), \modelname{} achieves the lowest error among the evaluated methods, suggesting improved stability under abrupt and localized disturbances.

\subsection{Parameter Sensitivity}
\begin{figure}[htbp]
    \centering
        \begin{subfigure}{0.33\textwidth}
        \centering
        \includegraphics[width=\textwidth]{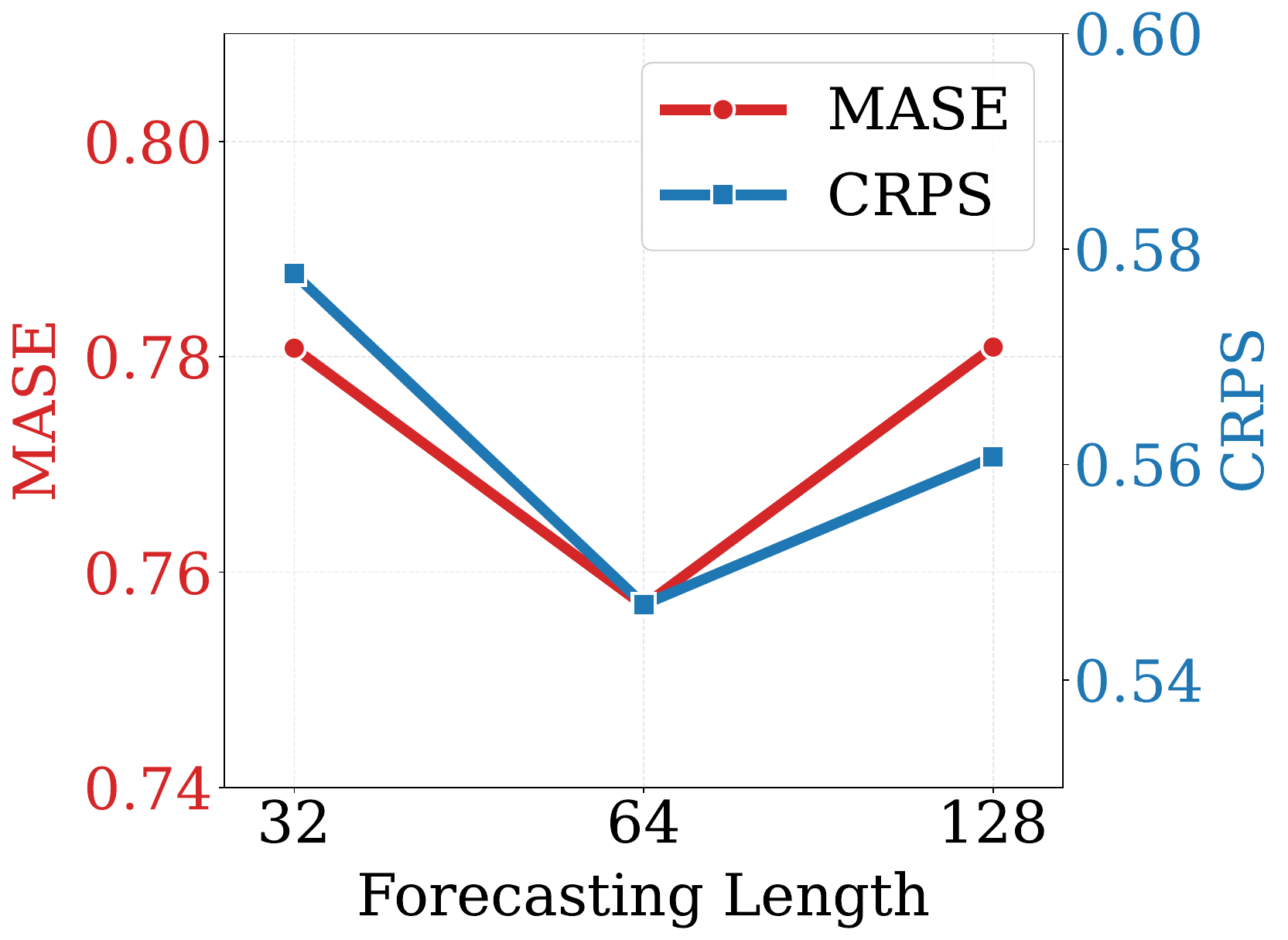}
        \caption{Forecasting length}
        \label{fig:param_fl}
    \end{subfigure}
    \begin{subfigure}{0.33\textwidth}
        \centering
        \includegraphics[width=\textwidth]{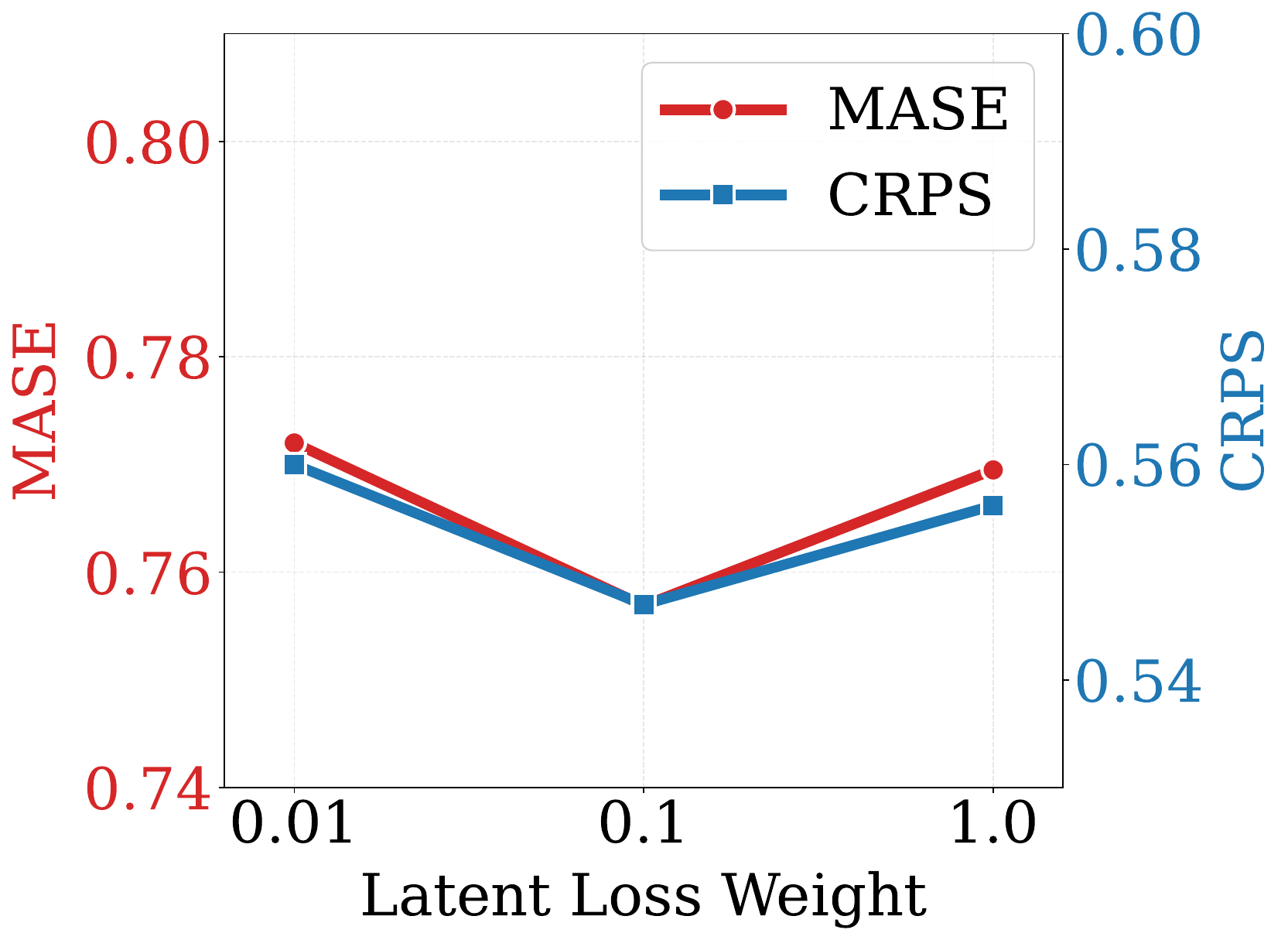}
        \caption{Latent loss weight}
        \label{fig:parms_lambda1}
    \end{subfigure}
    \begin{subfigure}{0.33\textwidth}
        \centering
        \includegraphics[width=\textwidth]{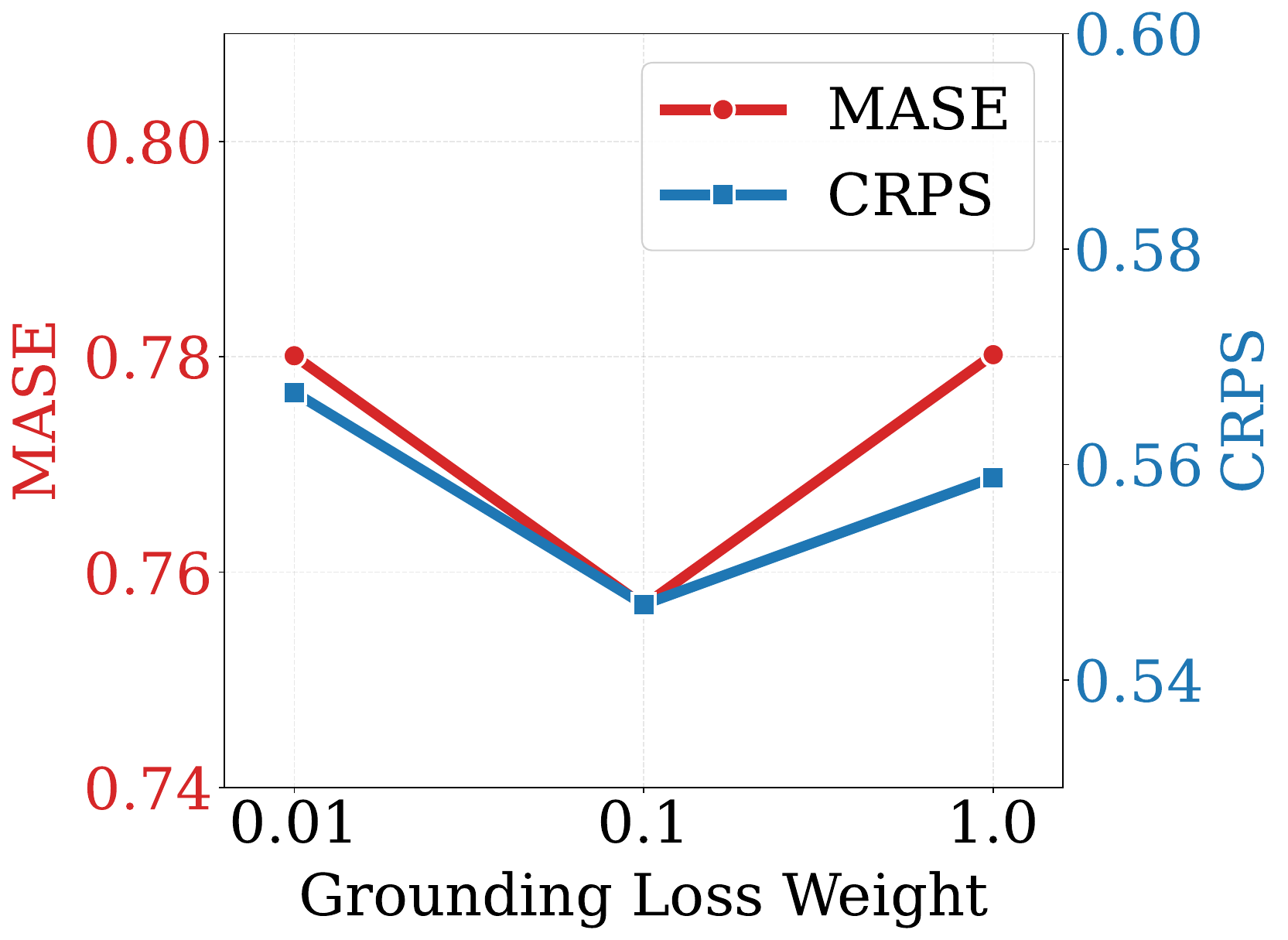}
        \caption{Grounding loss weight}
        \label{fig:parms_lambda2}
    \end{subfigure}
    \caption{Parameter sensitivity on the GIFT-Eval benchmark. Normalized MASE and CRPS under varying forecasting lengths and loss weights.}
    \label{fig:params}
\end{figure}

This section analyzes the sensitivity of the model to key hyperparameters, including the forecasting horizon for next-embedding prediction and the weights of the latent and grounding losses. For the forecasting horizon, we evaluate three settings $\{32, 64, 128\}$ on the GIFT-Eval benchmark, as shown in Figure~\ref{fig:param_fl}. Across these settings, the variations in MASE and CRPS are within approximately 0.2, indicating that the model is robust to the choice of forecasting horizon. Among the tested configurations, a forecasting horizon of 64 achieves the best overall performance. For the weights of the latent and grounding losses, we evaluate three scales $\{0.01,0.1,1\}$, as shown in Figures~\ref{fig:parms_lambda1} and~\ref{fig:parms_lambda2}. Across all weight settings, a weight of 0.1 yields the best overall performance, and the variations in MASE and CRPS remain within approximately 0.2, indicating low sensitivity to the loss weighting. The results suggest that the latent and grounding losses do not conflict with the forecasting objective; incorporating these losses leads to consistent convergence.

\subsection{Forecast Visualizations}
\label{app:forecast_visualizations}

\begin{figure}[htbp]
    \centering
        \begin{subfigure}{0.45\textwidth}
        \centering
        \includegraphics[width=\textwidth]{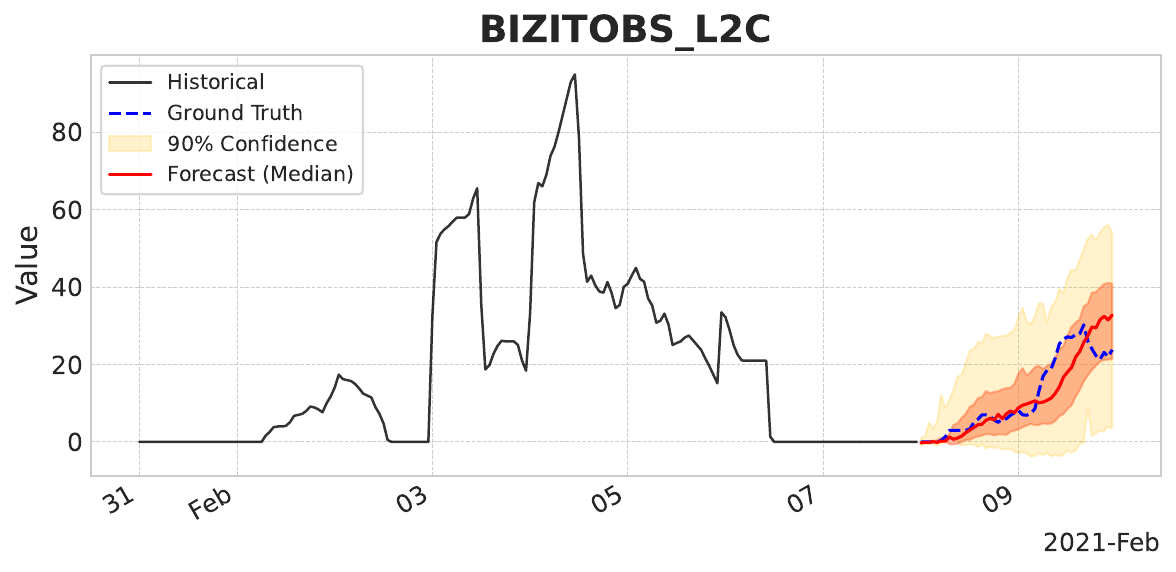}
    \end{subfigure}
    \begin{subfigure}{0.45\textwidth}
        \centering
        \includegraphics[width=\textwidth]{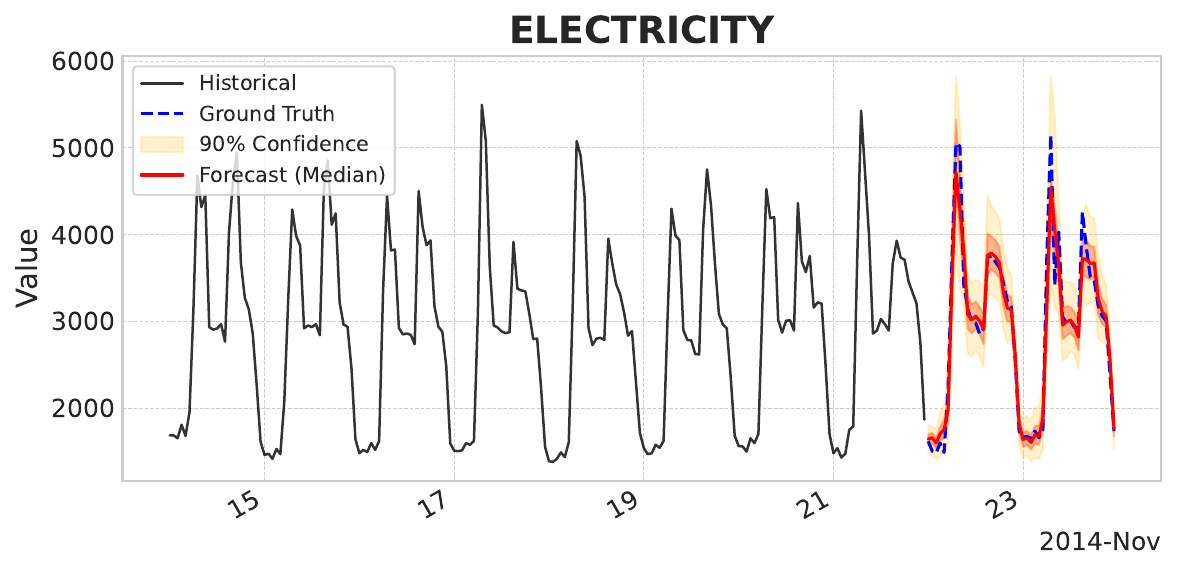}
    \end{subfigure}
    \centering
        \begin{subfigure}{0.45\textwidth}
        \centering
        \includegraphics[width=\textwidth]{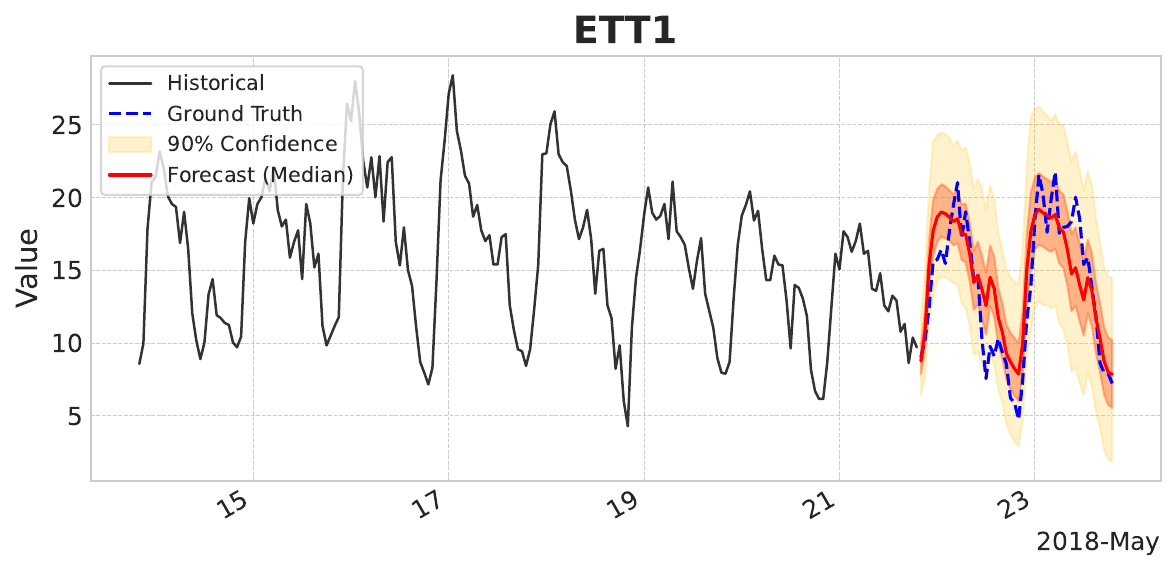}
    \end{subfigure}
    \begin{subfigure}{0.45\textwidth}
        \centering
        \includegraphics[width=\textwidth]{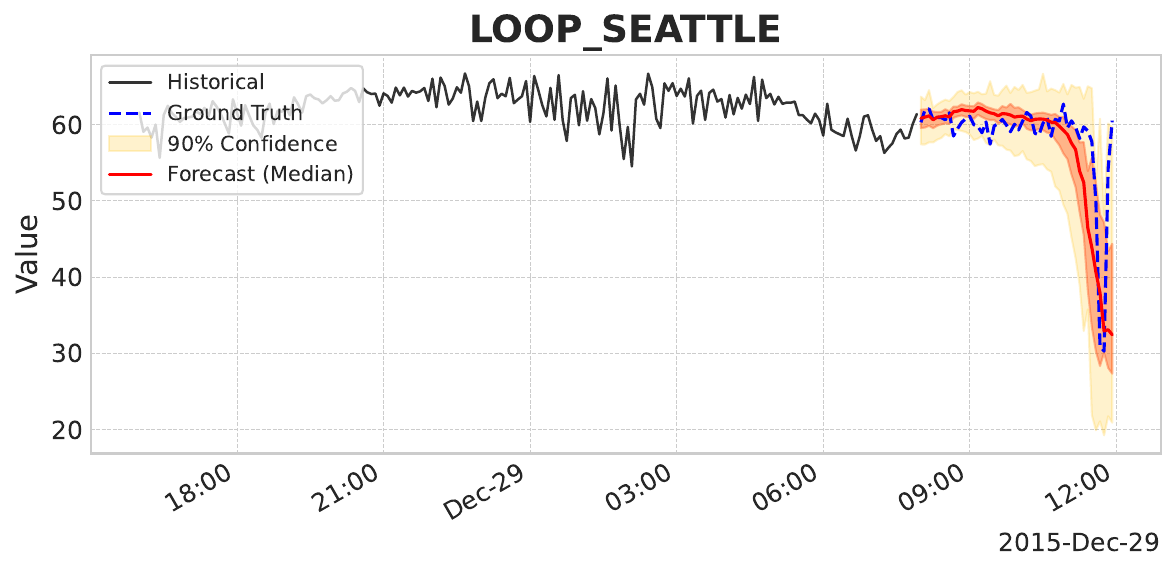}
    \end{subfigure}
    \centering
        \begin{subfigure}{0.45\textwidth}
        \centering
        \includegraphics[width=\textwidth]{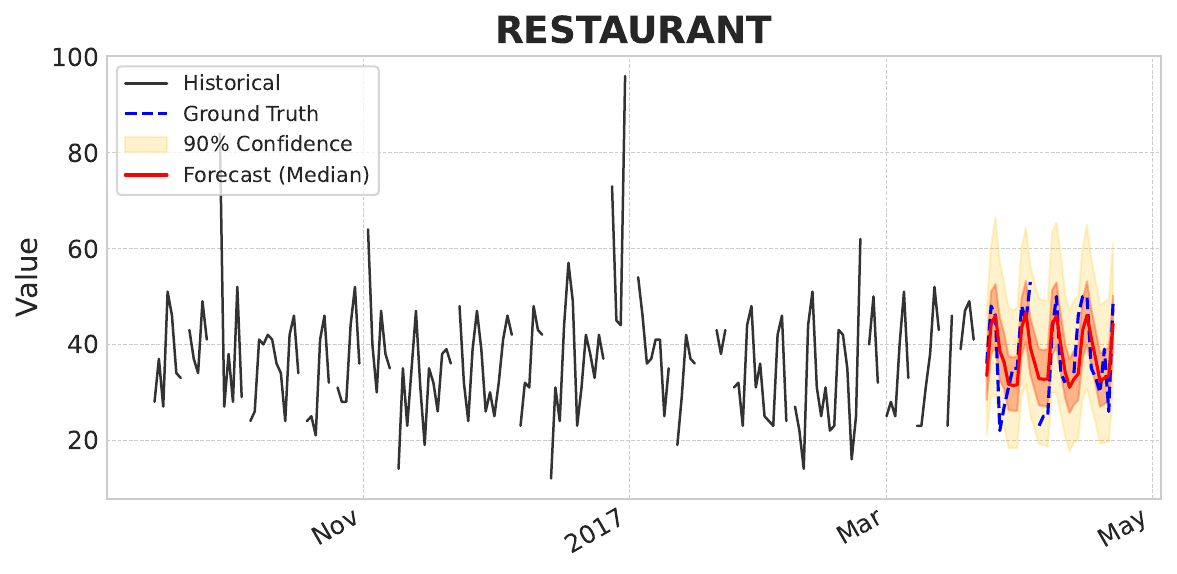}
    \end{subfigure}
    \begin{subfigure}{0.45\textwidth}
        \centering
        \includegraphics[width=\textwidth]{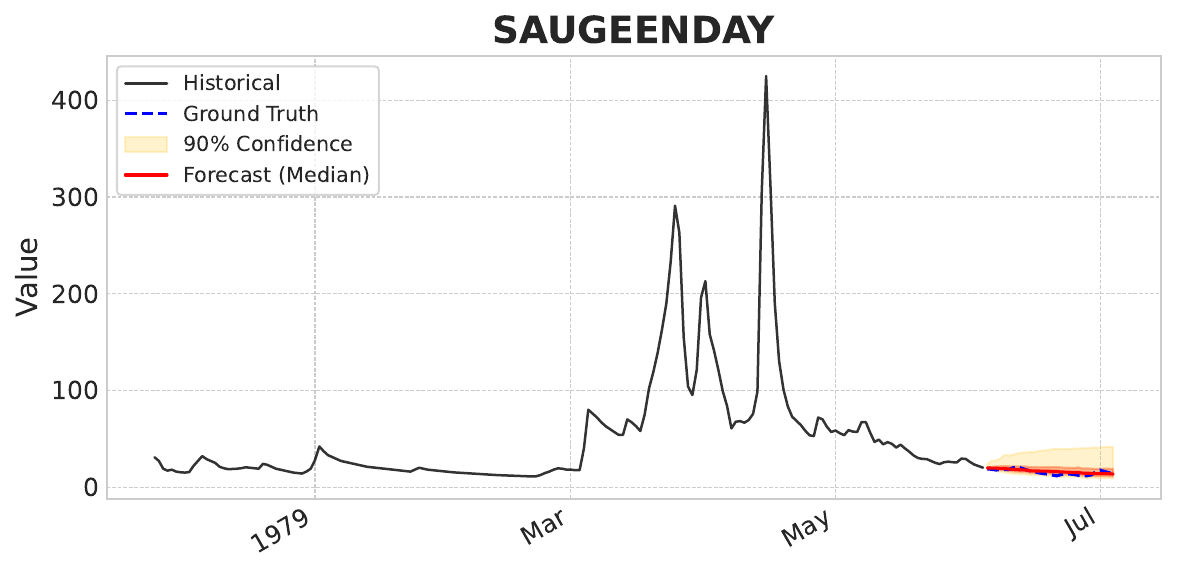}
    \end{subfigure}
    \centering
        \begin{subfigure}{0.45\textwidth}
        \centering
        \includegraphics[width=\textwidth]{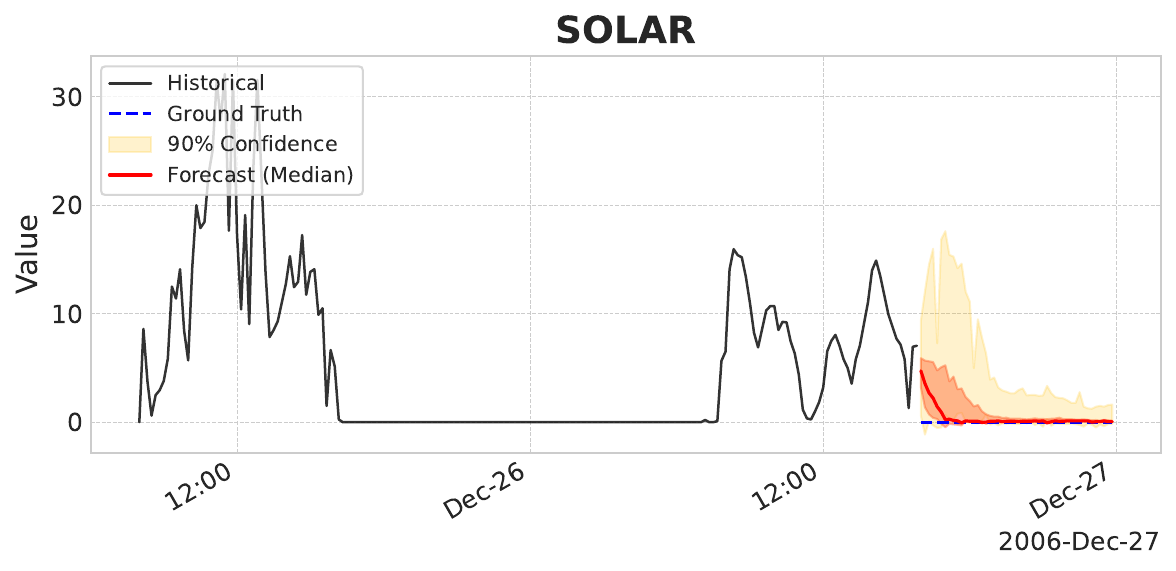}
    \end{subfigure}
    \begin{subfigure}{0.45\textwidth}
        \centering
        \includegraphics[width=\textwidth]{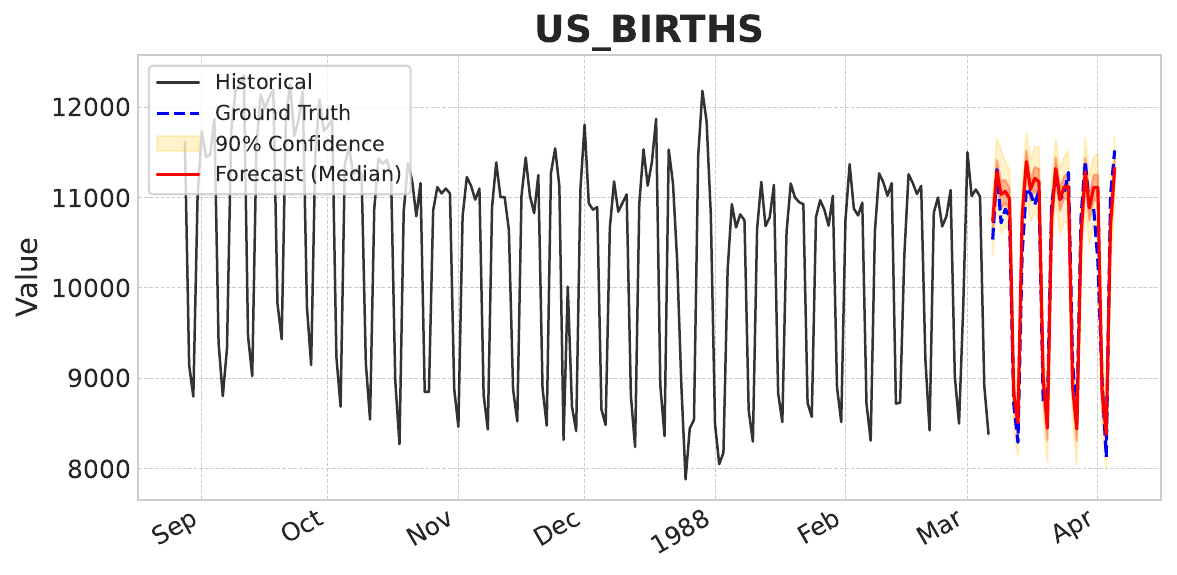}
    \end{subfigure}
    \centering
        \begin{subfigure}{0.45\textwidth}
        \centering
        \includegraphics[width=\textwidth]{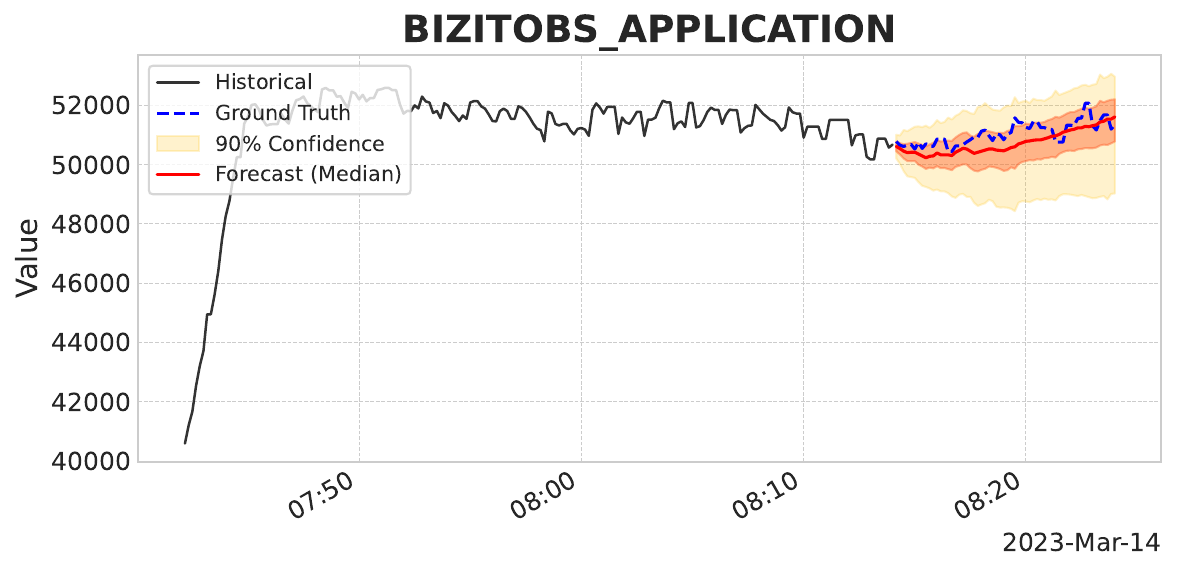}
    \end{subfigure}
    \begin{subfigure}{0.45\textwidth}
        \centering
        \includegraphics[width=\textwidth]{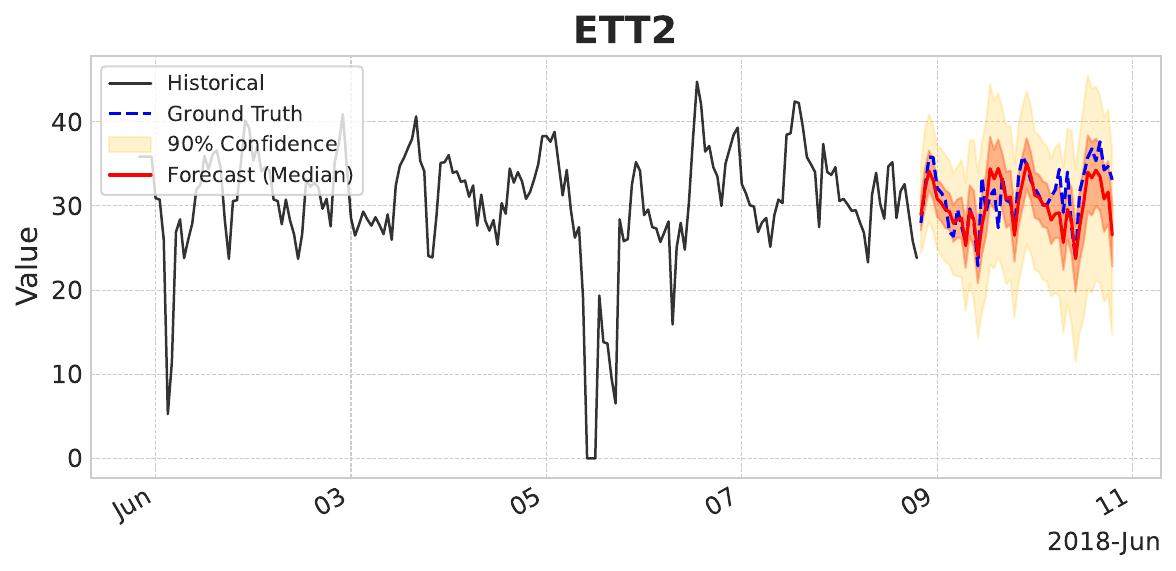}
    \end{subfigure}
    \centering
        \begin{subfigure}{0.45\textwidth}
        \centering
        \includegraphics[width=\textwidth]{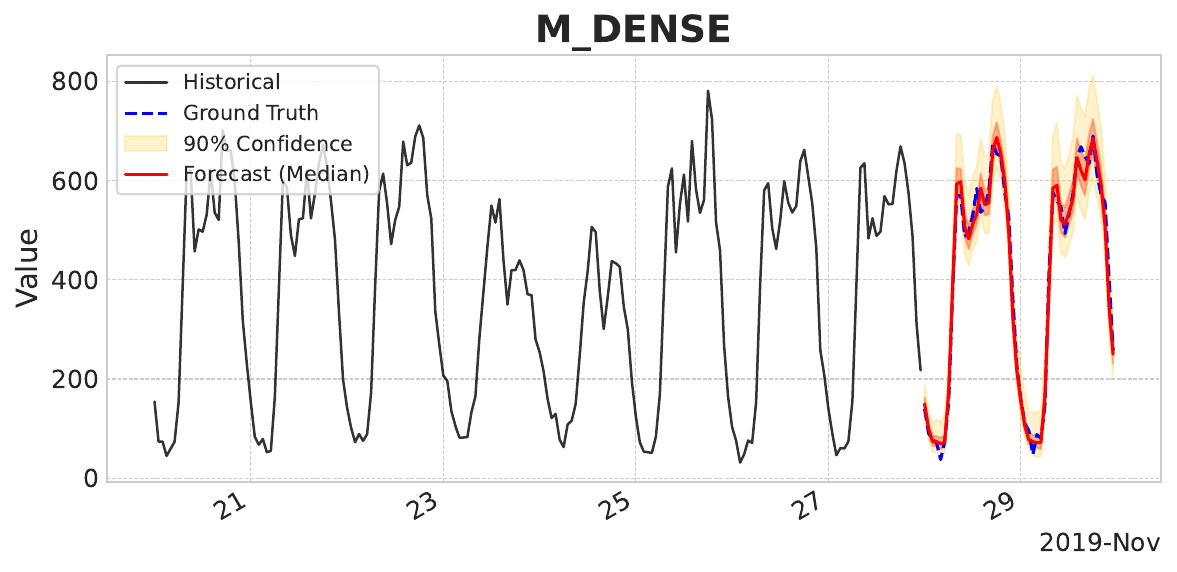}
    \end{subfigure}
    \begin{subfigure}{0.45\textwidth}
        \centering
        \includegraphics[width=\textwidth]{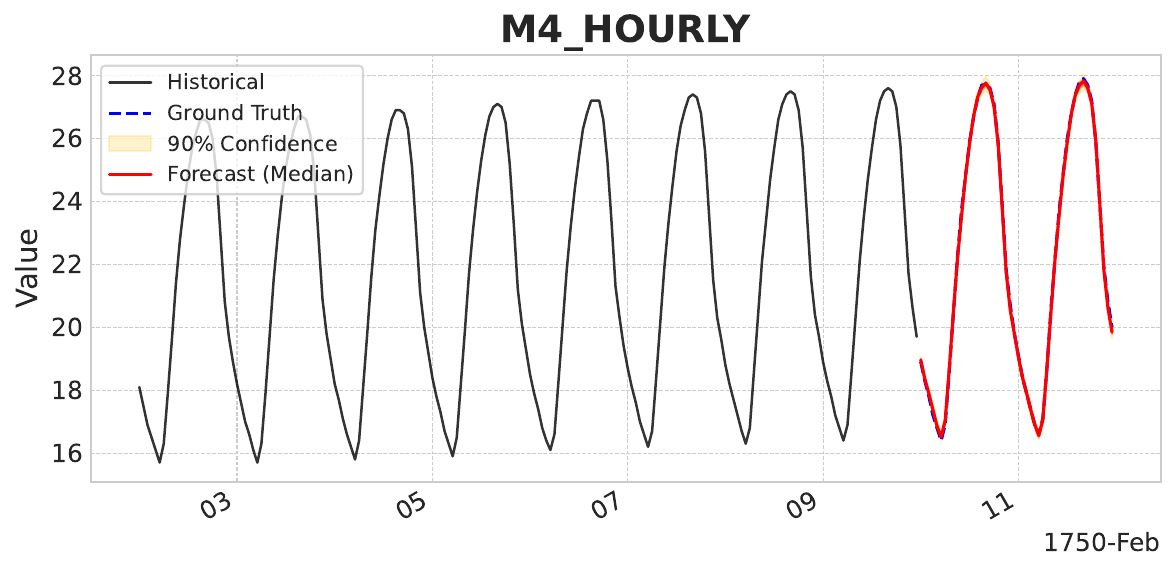}
    \end{subfigure}
    \caption{Showcases of zero-shot forecasts from \modelname{}.}
    \label{fig:forecast_visualizations}
\end{figure}

Figure~\ref{fig:forecast_visualizations} presents qualitative forecasting results produced by \modelname{} on representative datasets from the GIFT-Eval benchmark, covering both high-frequency series (e.g., traffic and energy-related data) and lower-frequency settings such as monthly births (US\_Births). All predictions are generated in a zero-shot setting given a fixed historical context window. The solid red line denotes the median forecast, while the shaded yellow region corresponds to the 90\% prediction interval. The visualizations illustrate the model’s ability to capture diverse temporal patterns, including sharp peaks in Electricity, smooth seasonal trends in M4\_Hourly, and intermittent behavior in Solar and Bizitobs\_L2C. The prediction intervals widen for series exhibiting higher volatility or irregular shifts, which is consistent with reasonable uncertainty quantification.

